\pdfoutput=1
\documentclass[10pt,twocolumn,letterpaper]{article}

\usepackage[pagenumbers]{cvpr} % To force page numbers, e.g. for an arXiv version

\usepackage{graphicx}
\usepackage{amsmath}
\usepackage{amssymb}
\usepackage{booktabs}
\usepackage{multirow}
\usepackage{comment}
\usepackage[pagebackref,breaklinks,colorlinks]{hyperref}

\usepackage{mysymbols}
\usepackage[capitalize]{cleveref}
\crefname{section}{Sec.}{Secs.}
\Crefname{section}{Section}{Sections}
\Crefname{table}{Table}{Tables}
\crefname{table}{Tab.}{Tabs.}

\def\cvprPaperID{5050} % *** Enter the CVPR Paper ID here
\def\confName{CVPR}
\def\confYear{2023}

\begin{document}

\title{Incremental 3D Semantic Scene Graph Prediction from RGB Sequences}
\author{Shun-Cheng Wu\textsuperscript{1}
\quad
Keisuke Tateno\textsuperscript{2}
\quad
Nassir Navab\textsuperscript{1}
\quad
Federico Tombari\textsuperscript{1,2}\\
\textsuperscript{1}Technische Universit\"{a}t M\"{u}nchen \hspace{1.5cm} \textsuperscript{2}Google\\[0.1cm]
%Boltzmannstraße 3, 85748, Germany. Brandschenkestrasse 110, 8002 Zürich\\
% {\small shuncheng.wu, johanna.wald, nassir.navab@tum.de, ktateno, tombari@google.com}}
% {\small shuncheng.wu@tum.de}
% {\small \href{https://shunchengwu.github.io/MonoSSG}{shunchengwu.github.io/MonoSSG}}
}
\maketitle

%%%%%%%%% ABSTRACT
\begin{abstract}
% briefly narrow down the topic
3D semantic scene graphs are a powerful holistic representation as they describe the individual objects and depict the relation between them. They are compact high-level graphs that enable many tasks requiring scene reasoning. In real-world settings, existing 3D estimation methods produce robust predictions that mostly rely on dense inputs. In this work, we propose a real-time framework that incrementally builds a consistent 3D semantic scene graph of a scene given an RGB image sequence. Our method consists of a novel incremental entity estimation pipeline and a scene graph prediction network. The proposed pipeline simultaneously reconstructs a sparse point map and fuses entity estimation from the input images. The proposed network estimates 3D semantic scene graphs with iterative message passing using multi-view and geometric features extracted from the scene entities. Extensive experiments on the 3RScan dataset show the effectiveness of the proposed method in this challenging task, outperforming state-of-the-art approaches. %
Our implementation is available at \url{https://shunchengwu.github.io/MonoSSG}.
\end{abstract}

%%%%%%%%% BODY TEXT
\begin{figure}[t]
    \centering
    %\resizebox{\columnwidth}{!}{
    \includegraphics[width=\columnwidth]{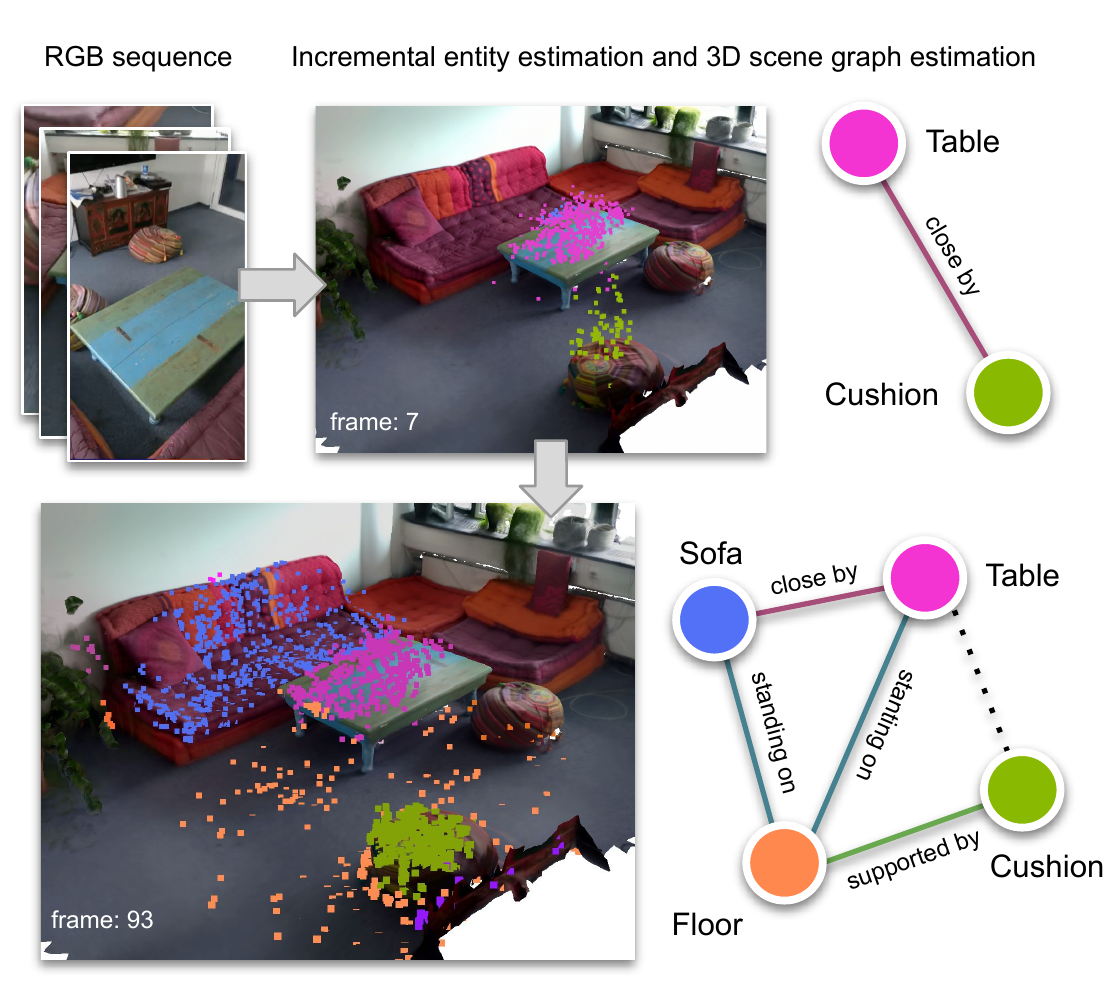}
    %}
    \caption{We propose a real-time 3D semantic scene graph estimation method that relies on an abstract understanding of a scene geometry built with RGB input. Our method estimates scene graphs incrementally by continuously estimating scene graphs and fusing local predictions into a global 3D scene graph.}
    \label{fig:teaser}
    \vspace{-0.5cm}
\end{figure}%
\section{Introduction}\label{sec:intro}
%--- structure ---
\begin{comment}
\begin{itemize}
    \item general intro
    \begin{itemize}
        \item Scene understanding
        \item Scene Graphs
        \item 3D Scene graphs
        \item Incremental 3D Scene Graphs
    \end{itemize}
    \item difficulties
    \begin{itemize}
        \item Exact 3D shape is not always available.
        \item 
    \end{itemize}
    \item ours
    \item summary
\end{itemize}
\end{comment}
% ========
% General Intro
% ========
% ---Scene understanding is a fundamental block. ---
Scene understanding is a cornerstone in many computer vision applications requiring perception, interaction, and manipulation, such as robotics, AR/VR and autonomous systems~\cite{stasse2006real,scona2017direct,grotz2017graph,seiwald2021lola}. Semantic Scene Graphs (SSGs) go beyond recognizing individual entities (objects and stuff) by reasoning about the relationships among them~\cite{Xu2017,Wald2020_3dssg}. They also proved to be a valuable representation for complex scene understanding tasks, such as image captioning~\cite{Xu2015,Karpathy2015}, generation\cite{Johnson2018,garg2021unconditional}, scene manipulation\cite{dhamo2020semantic,dhamo2021graph}, task planning~\cite{kim20193}, and surgical procedure estimation~\cite{Ozsoy2022_4D_OR,Ozsoy2021_MSSG}. Given the benefits of such representations, scene graph estimation received increasing attention in the computer vision community.\par
While earlier methods mainly estimate SSGs from images~\cite{gu2019unpaired,gu2019scene,zhong2021learning,liu2021fully,Xu2017}, recent approaches have also investigated estimating them from 3D data. Compared to 2D scene graphs, which describe a single image, 3D scene graphs depict the entire 3D scenes, enabling applications requiring a holistic understanding of the whole scene, such as path planning~\cite{ravichandran2022hierarchical}, camera localization, and loop closure detection~\cite{hughes2022hydra}. %
However, existing 3D methods either require dense 3D geometry of the scenes to estimate 3D scene graphs~\cite{Armeni2019_3dsg,Wald2020_3dssg,wu2021scenegraphfusion,hughes2022hydra}, which limits the use case since dense geometry is not always available, or constraints the scene graph estimation at the image-level~\cite{Xu2017,Gay2018_visual,kim20193}, which tend to fail inferring relationships among objects beyond the individual viewpoints. A method that estimates 3D scene graphs relies on sparse scene geometry and reasoning about relationships globally has not been explored yet.\par%
%
% ---we propose---
In this work, we propose a real-time framework that incrementally estimates a global 3D SSG of a scene simply requiring an RGB sequence as input. The process is illustrated in \FIG{teaser}. Our method simultaneously reconstructs a segmented point cloud while estimating the SSGs of the current map. The estimations are bound to the point map, which allows us to fuse them into a consistent global scene graph. The segmented map is constructed by fusing entity estimation from images to the points estimated from a sparse Simultaneous Localization and Mapping (SLAM) method~\cite{campos2021orb3}. Our network takes the entities and other properties extracted from the segmented map to estimate 3D scene graphs. %
Fusing entities across frames is non-trivial. Existing methods often rely on dense inputs~\cite{Tateno2015,Gaku2019_panopticfusion} and struggle with sparse inputs since the points are not uniformly distributed. Estimating scene graphs with sparse input points is also challenging. Sparse and ambiguous geometry renders the node representations unreliable. On the other hand, directly estimating scene graphs from 2D images ignores the relationship beyond visible viewpoints. %
We aim to overcome the aforementioned issues by proposing two novel approaches. First, we propose a confidence-based fusion scheme which is robust to variations in the point distribution. Second, we present a scene graph prediction network that mainly relies on multi-view images as the node feature representation. Our approach overcomes the need for exact 3D geometry and is able to estimate relationships without view constraints. %
In addition, our network is flexible and generalizable as it works not only with sparse inputs but also with dense geometry.\par%
%
% ---summary---
We comprehensively evaluate our method on the 3D SSG estimation task from the public 3RScan dataset~\cite{Wald2019RIO}.
We experiment and compare with three input types, as well as 2D and 3D approaches. Moreover, we provide a detailed ablation study on the proposed network. The results show that our method outperforms all existing approaches by a significant margin.
The main contributions of this work can be summarized as follows:
(1)~We propose the first incremental 3D scene graph prediction method using only RGB images. %
(2)~We introduce an entity label association method that works on sparse point maps. %
(3)~We propose a novel network architecture that generalizes with different input types and outperforms all existing methods.%
\section{Related Work}
\subsection{3D Object Localization from Images}%
%onlien v.s. Offline
%semantic v.s. Instance aware
%with the semantic scene graph.
%sparse v.s. dense
Localizing 3D objects from images aims to predict the position and orientation of objects. Existing methods can be broadly divided into two categories: without and with explicit geometrical reasoning.\par%%
In the former category, many works focus on estimating 3D bounding boxes by extending 2D detectors with learned priors~\cite{li2018unified,mousavian20173d,nie2020total3dunderstanding,zhang2021holistic}. When sequential input is available, single view estimations can be fused to estimate a consistent object map~\cite{brazil2020kinematic,hu2019joint,li2021odam,li2021moltr}. However, the fused results may not fulfill the multi-view geometric constraints. Multi-view approaches estimate oriented 3D bounding boxes from the given 2D detection of views. They mainly focus on minimizing the discrepancies between the projected 3D representation and the detected 2D bounding boxes.\par%%
In the latter category, 3D objects are localized with the help of explicit geometric information. Many existing methods treat object detection as spatial landmarks in a map~\cite{hosseinzadeh2018structure,McCormac2018_fusionplusplus,nicholson2018quadricslam,SalasMoreno2013,wu2020eao,yang2019cubeslam}, also known as object-level SLAM. Others focus on fusing dense per-pixel predictions to a reconstructed map~\cite{McCormac2017_semanticfusion,chen2019suma_plusplus,Gaku2019_panopticfusion,Jiazhao2020_fusionaware,pham2019real,Grinvald2019}, which is known as semantic mapping or semantic SLAM.\par%
A major difference between object-level and semantic SLAM is that the former focuses only on foreground objects, while the latter also considers the structural and background information. %
Specifically, SemanticFusion~\cite{McCormac2017_semanticfusion} fuses dense semantic segments from images to a consistent dense 3D map with Bayesian updates. Its map representation provides a dense semantic understanding of a scene ignoring individual instances. %
PanopticFusion~\cite{Gaku2019_panopticfusion} proposes to combine the instance and semantic segmentation from images to a panoptic map. Their approach considers foreground object instances and non-instance semantic information from the background. SceneGraphFusion~\cite{wu2021scenegraphfusion} relies on 3D geometric segmentation~\cite{tateno2017large} and scene graph reasoning to achieve instance understanding of all entities in a scene.\par%
One significant difficulty in instance estimation for semantic SLAM is associating the instances across frames. Existing approaches mainly rely on a dense map to associate predictions by calculating the intersection-over-union (IoU) or the overlapping ratio between the input and the rendered image from the map. However, these methods produce sub-optimal results when the map representation is sparse due to the non-uniform distribution of the map points. 
We overcome this problem by proposing a confidence-based association.\par%
%
% https://docs.google.com/drawings/d/1qRSXUTHRbdU9XeYeYuQ2mflHglZgQUzGWvn1b7QUxfg/edit
\begin{figure*}[t]
    \centering
    \includegraphics[width=\textwidth]{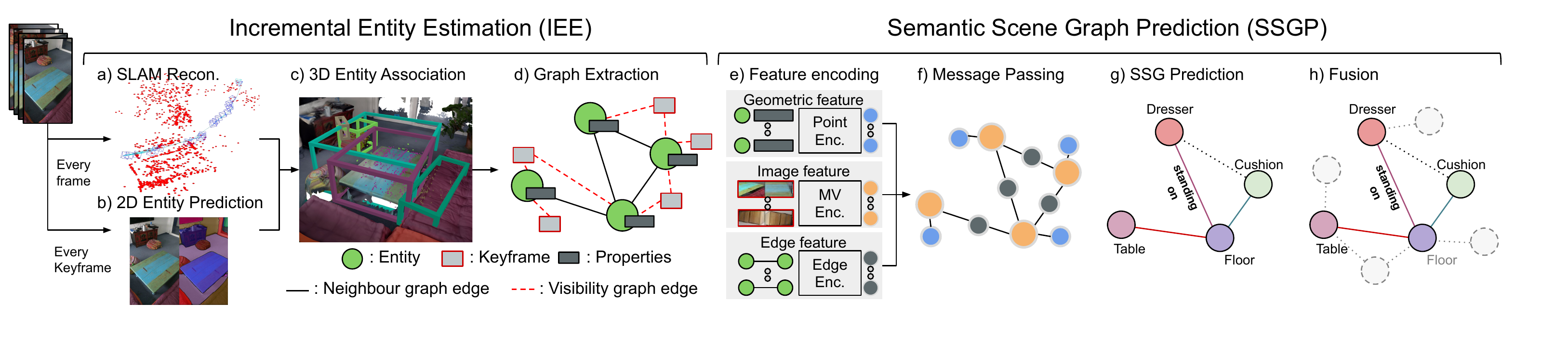}
    % \vspace{-0.5cm}
    \caption{
    Given a sequence of RGB images, we use every frame to reconstruct a sparse point map (a) and every keyframe to estimate the 2D entities (b). (a) and (b) are associated and merged into a single 3D map (c). We asynchronously extract graph properties from the entity map (d) to estimate SSG. Our network computes the geometric, multi-view, and edge features (e). These features are propagated to each other with message passing (f), then used to predict a SSG (g). Then, periodic SSGs are fused to a global 3D SSG (h).
    }\label{fig:sys_pipeline}
    % \vspace{-0.5cm}
\end{figure*}%%
\subsection{3D Semantic Scene Graph}%
% The list of the most related works %
% IMP~\cite{Xu2017}: 2D
% VGfM~\cite{Gay2018_visual}: 2D+consistency RGB
% 3D-S-G~\cite{kim20193}: 2D + consistency RGBD
% Armeni~\cite{Armeni2019_3dsg}: 3D offline
% 3D Dynamic Scene~\cite{Rosinol2020_dynamic}: 3D offline
% Kimera + 3D Dynamic Scene~\cite{rosinol2021kimera}
% 3DSSG~\cite{Wald2020_3dssg}: 3D offline
% SGFN~\cite{wu2021scenegraphfusion}: 3D online RGBD
% Hydra~\cite{hughes2022hydra}: 3D online RGBD
Estimating 3D scene graphs methods can be divided according to various criteria. From the perspective of the scene graph structure, some methods focus on hierarchical scene graphs~\cite{Armeni2019_3dsg,Rosinol2020_dynamic,rosinol2021kimera,hughes2022hydra}. These approaches mainly address the problem of relationship estimation between entities from different hierarchical levels, \eg~object, and room level. The other methods focuses on pairwise relationships, \eg~support and comparative relationships, between nodes within a scene~\cite{Wald2020_3dssg,kim20193,Xu2017,wu2021scenegraphfusion}. %
From the perspective of input data, some previous methods rely on RGB input~\cite{Gay2018_visual,kim20193} by fusing 2D scene graph predictions to a consistent 3D map. On the other hands, other methods rely on 3D input~\cite{Armeni2019_3dsg,Rosinol2020_dynamic,Wald2020_3dssg,rosinol2021kimera,wu2021scenegraphfusion,hughes2022hydra} by using known 3D geometries. Nevertheless, most of the existing methods estimate scene graphs offline~\cite{Gay2018_visual,Armeni2019_3dsg,rosinol20203d,Rosinol2020_dynamic}, while a few works~\cite{kim20193,wu2021scenegraphfusion,hughes2022hydra} predict scene graph in real-time.\par
Among all existing work, the pioneering work in 3D scene graph estimation is proposed by \cite{Gay2018_visual}. The authors extend the 2D scene graph estimation method from~\cite{Xu2017} with temporal consistency across frames and use geometric features from ellipsoids. Kim \etal~\cite{kim20193} propose an incremental framework to estimate 3D scene graphs from 2D estimations. Armeni \etal~\cite{Armeni2019_3dsg} are the first work to estimate 3D scene graphs through the hierarchical understanding of the scene. Rosinol \etal~\cite{Rosinol2020_dynamic} build on top of~\cite{Armeni2019_3dsg} to capture moving agents. This work is subsequently extended to a SLAM system~\cite{rosinol2021kimera}.
Wald \etal~\cite{Wald2020_3dssg} propose the first 3D scene graph method based on the relationship between objects at the same level, along with 3RScan: a richly annotated 3D scene graph dataset. Wu \etal~\cite{wu2021scenegraphfusion} extend~\cite{Wald2020_3dssg} to real-time scene graph estimation with a novel feature aggregation mechanism. Hughes \etal~\cite{hughes2022hydra} propose to reconstruct a 3D hierarchical scene graph in real-time incrementally. Our method incrementally estimates a flat scene graph with multi-view RGB input and a sparse 3D geometry, which differentiates our work from the previous methods relying on 3D input~\cite{Wald2020_3dssg,wu2021scenegraphfusion}, and approaches without geometric understanding~\cite{Gay2018_visual,kim20193}.
%%%%%%%%%%%%%%%%%%%%%%%%%%%%%%%%%%%%%%%%%%%%%%%%%%%%%%%%%%%%%%%%%%%%%%%%%%%%%%%%%%%%%%%%%%%%%%%%%%%%%%%%%%%%%%
%%%%%%%%%%%%%%%%%%%%%%%%%%%%%%%%%%%%%%%%%%%%%%%%%%%%%%%%%%%%%%%%%%%%%%%%%%%%%%%%%%%%%%%%%%%%%%%%%%%%%%%%%%%%%%
%
\section{Method}%3D Scene Graph from RGB images
% The proposed pipeline aims to estimate 3D scene graphs from RGB sequences incrementally. \FIG{sys_pipeline} illustrates the overview of our proposed pipeline. 
The proposed framework is illustrated in \FIG{sys_pipeline}, which shows how, given a sequence of RGB images, it can estimate a 3D semantic scene graph incrementally. The Incremental Entity Estimation (IEE) front end makes use of the images to generate segmented sparse points. Those are merged into 3D entities and used to generate both an entity visibility graph and a neighbour graph. The Semantic Scene Graph Prediction (SSGP) network uses the entities and both graphs to estimate multiple scene graphs and then fuse them into a consistent 3D SSG.\par%
We define a SSG as $\sSceneGraph=\left(\sNodes,\sEdges\right)$, where $\sNodes$ and $\sEdges$ denote a set of entity nodes and directed edges. %
Each node $\sNode_i\in\sNodes$ is assigned an entity label $\sLabel_{i}\in\sLabels$, a set of points $\sPoints_i$, an Oriented Bounding Box (OBB) $\sBBox_{i}$ and a node category $\sNodeCls_{i}\in\sNodesCls$. Each edge $\sEdge_{i\to j}\in\sEdges$, connecting node $\sNode_i$ to $\sNode_j$ where $i\neq j$, consists of an edge category $\sEdgeCls_{i\to j}\in\sEdgesCls$. $\sLabels$, $\sNodesCls$, and $\sEdgesCls$ denote all entity labels, a node category set, and an edge category set, respectively. %
An OBB $\sBBox_{i}$ is a gravity-aligned 3D bounding box consisting of a boundary dimension $\sBBoxSize_{i}\in\mathbb{R}^{3}$, a center $\sCenter_{i}\in\mathbb{R}^{3}$, and an angle that encodes the rotation along the gravity axis. The OBBs are used to build both graphs and features. %, and is used to build part of node and edge features.
The entity visibility graph models the visibility relationship of the entities as a bipartite graph $\sCoVisGraph = (\sNodes, \sKFs, \sEdges_c)$ where $\sKFs$, $\sEdges_c$ denote a set of keyframes and visibility edges, respectively. \(\sCoVisGraph\) gives the knowledge of the visibility of entity nodes in keyframes, which is used in computing multi-view visual features in SSGP. %
The neighbour graph encodes the proximity relationship of the entities as an undirected graph $\sProxGraph = (\sNodes, \sEdges_p)$, where \(\sEdges_p\) is the set of proximity edges. The neighbour graph also serves as the initial graph for the message propagation step in SSGP.\par%
%
% We will first introduce how we estimate entities and graphs from RGB images (\SEC{IEE}), and how those estimations are used to estimate a scene graph (\SEC{network}).\par%
%%%%%%%%%%%%%%%%%%%%%%%%%%%%%%%%%%%%%%%%%%%%%%%%%%%%%%%%%%%%%%%%%%%%%%%%%%%%%%%%%%%%%%%%%%%%%%%%%%%%%%%%%%%%%%%%%
%%%%%%%%%%%%%%%%%%%%%%%%%%%%%%%%%%%%%%%%%%%%%%%%%%%%%%%%%%%%%%%%%%%%%%%%%%%%%%%%%%%%%%%%%%%%%%%%%%%%%%%%%%%%%%%%
\subsection{Incremental Entity Estimation}\label{sec:IEE}%
During the first step of the IEE front end pipeline, a set of labeled 3D points are estimated from the sequence of RGB images (\SEC{slam}). The entity labels are determined using an entity segmentation method on selected keyframes (\SEC{entity_detection}). Then, they are associated and fused into a sparse point map (\SEC{entity_fusion}). Finally, the entities and their properties are extracted using the labeled 3D points (\SEC{property_extraction}).\par%
%
%%%%%%%%%%%%%%%%%%%%%%%%%%%%%%%%%%%%%%%%%%%%%%%%%%%%%%%%%%%%%%%%%%%%%%%%%%%%%%%%%%%%%%%%%%%
\subsubsection{Sparse Point Mapping}\label{sec:slam}%
We use ORB-SLAM3~\cite{campos2021orb3} to simultaneously estimate the camera poses and build a sparse point map by matching estimated keypoints from sequential RGB frames. To guarantee real-time performance, an independent thread is used to run the local mapping process using the stored keyframes. The same thread additionally takes care of running the entity detector and performing the label mapping process. For each point $\sPoint_m\in\sAllPoints$ in the map, we store its 3D coordinates, an entity label $\sLabel_{m}$, and its confidence score $\sWeight_{m}\in\mathbb{R}_{\geq 0}$.\par%
%
%%%%%%%%%%%%%%%%%%%%%%%%%%%%%%%%%%%%%%%%%%%%%%%%%%%%%%%%%%%%%%%%%%%%%%%%%%%%%%%%%%%%%%%%%%%
\subsubsection{2D Entity Detection}\label{sec:entity_detection}%
We estimate an entity label mask $\sImgMasks_{t}(\sImgCoord)\in \sImgLabels_{t}$ and a confidence mask $\sImgWeights_{t}(\sImgCoord) \in [0,1]\subset \mathbb{R}$ with every given keyframe $\sKF_t\in\sKFs$, where $\sImgCoord\in\mathbb{R}^{2}$ denotes the image coordinates and $\sImgLabels_{t}$ all entity labels in $\sKF_t$. Both masks are estimated using a class-agnostic segmentation network which further improves other instance segmentation methods~\cite{carion2020end,cheng2021per,li2021fully} by enabling the discovery of unseen entities~\cite{qi2021entity,joseph2021towards,gasperini2022holistic}. Although segmentation networks provide accurate masks, the estimations are independent across frames. Thus, a label association stage is required to build a consistent label map.\par%
%
%%%%%%%%%%%%%%%%%%%%%%%%%%%%%%%%%%%%%%%%%%%%%%%%%%%%%%%%%%%%%%%%%%%%%%%%%%%%%%%%%%%%%%%%%%%
\subsubsection{Label Association and Fusion}\label{sec:entity_fusion}%
Inspired by~\cite{Tateno2015,runz2018maskfusion,McCormac2018_fusionplusplus,Gaku2019_panopticfusion}, we use the reference map approach to handle the label inconsistency. It relies on a map reconstruction to solve label consistency by comparing input label mask to rendered mask. Then fuse the associated mask to the global point map.\par%
%
% \noindent\textbf{Label Association.}
\noindent\textbf{Label Association.}
We start by building a reference entity mask $\sRefMasks_{t}(\sImgCoord)\in\sLabels$ by projecting point entity labels from the sparse point map using the pose of $k_{t}$. The consistency-resolved entity mask $\sConsistImgMasks_{t}(\sImgCoord)$ is estimated by evaluating the corresponding labels on the image mask $\sImgMasks_{t}(\sImgCoord)$ and the reference mask $\sRefMasks_{t}(\sImgCoord)$. 
This evaluation can be performed by different methods such as using intersection over union statistics~\cite{Gaku2019_panopticfusion} or the maximum overlapping ratio between label masks~\cite{Tateno2015}. However, both methods assume that points are uniformly distributed; a premise that fails in most sparse point reconstruction tasks. In such cases, these methods become unstable, as shown in the example provided in the supplementary material.
To overcome this problem, we propose to use the maximum mean confidence as the criteria to find the best candidate. First, a confidence mask $\sRefWeights_{t}(\sImgCoord)$ is built by projecting the point label confidence using the pose of $k_{t}$, then the mean confidence score of a label $\sImgLabel\in\sImgMasks_{t}(\sImgCoord)$ and a reference label $\sRefLabel\in\sRefMasks_{t}(\sImgCoord)$ is computed by
\begin{equation}\label{eq:mean_confidence_calculation}
\funcMCS{\sImgLabel}{\sRefLabel} = \frac{\sum_{u' \in \funcCorresSet{\sImgLabel}{\sRefLabel}} \sRefWeights_{t}(u')} {\funcCardi{\funcCorresSet{\sImgLabel}{\sRefLabel}}},
\end{equation}
where $\funcCorresSet{\sImgLabel}{\sRefLabel}$ gives a set of image coordinates $u'\in\mathbb{R}^{2}$ where $\sImgLabel$ and $\sRefLabel$ overlap: $\{ u'\mid ({\sImgMasks_{t}(u')=\sImgLabel})\land ( {\sRefMasks_{t}(u')=\sRefLabel} ) \}$, and $\funcCardi{\cdot}$ is the cardinality operator. %
Then, the mask $\sConsistImgMasks_{t}(\sImgCoord)$ is generated by replacing the per-pixel entity label $\sImgLabel\in\sImgMasks_{t}(\sImgCoord)$ with either a reference label $\sRefLabel$ or a new label $\sLabel_{\text{new}} \notin \sLabels$ depending on:
\begin{equation}\label{eq:label_association_new}
\sImgLabel = 
\begin{cases}
    {\arg\max}_{\sRefLabel}~\funcMCS{\sImgLabel}{\sRefLabel} & \text{if }
    {\max}_{\sRefLabel}~{\frac{\funcCardi{\funcCorresSet{\sImgLabel}{\sRefLabel}}}{\funcCardi{\sRefMasks_{t}(\sImgCoord)=\sRefLabel}}}>\sIoUThres\\
    % {\max}_{\sRefLabel}~\funcCorresSetScaled{\sImgLabel}{\sRefLabel}>\sIoUThres\\
    \sLabel_{\text{new}} &\text{otherwise}
\end{cases}
,
\end{equation}%
where we filter out a match if the number of overlapped pixel has a low coverage over the total number of label $\sRefLabel$ on the reference mask with a threshold $\sIoUThres$. %
% Here $\funcCorresSetScaled{\sImgLabel}{\sRefLabel}$ is $\funcCardi{\funcCorresSet{\sImgLabel}{\sRefLabel}}$ normalized by the total number of label $\sRefLabel$ on the reference mask, \ie %
% $\funcCorresSetScaled{\sImgLabel}{\sRefLabel}=\frac{\funcCardi{\funcCorresSet{\sImgLabel}{\sRefLabel}}}{\funcCardi{\sRefMasks_{t}(\sImgCoord)=\sRefLabel}}$, and $\sIoUThres$ is a threshold to filter the correspondences with insufficient corresponding points. %
%
In addition, similar to \cite{Gaku2019_panopticfusion}, a reference entity label is assigned to only one input entity label. If a reference label has been assigned, we use descending order to search for the next best candidate.\par% based on the mean confidence score if the current one has been assigned.\par%
%
%%% FUSION %%%
% \noindent\textbf{Label Fusion.}
\noindent\textbf{Label Fusion.}
After the association process, the associated entity labels $\sConsistImgMasks_{t}(\sImgCoord)$ are fused to the sparse point map $\sAllPoints$. 
Since each label on $\sConsistImgMasks_{t}(\sImgCoord)$ sources from a map point, the label and confidence value of a point are updated by
% This is achieved by integrating label confidence to the corresponding points on the mask
% 
% Since the reference mask is generated by projecting map point, we update the label and confidence of a point based on the mask.
% 
% We point index on which points are projected to the masks are tracked, which allows us to 
% 
% Therefore, we update entity label and confidence of the map points that are projected on the reference masks $\sRefMasks_{t}(\sImgCoord)$ and $\sRefWeights_{t}(\sImgCoord)$. %
%
% Given a function $\psi(\sImgCoord)$ that returns the corresponding point index that is projected on the pixel location $u$ on both $\sRefMasks_{t}(\sImgCoord)$ and $\sRefWeights_{t}(\sImgCoord)$. 
% 
% We update the entity confidence on map points by 
\begin{equation}
\sWeight_{\psi\left(\sImgCoord\right)}=
\begin{cases}
\sWeight_{\psi\left(\sImgCoord\right)}+\sImgWeights_{t}\left(\sImgCoord\right)  & \text{if }\sRefMasks_{t}\left(\sImgCoord\right)=\sConsistImgMasks_{t}\left(\sImgCoord\right)\\
\sWeight_{\psi\left(\sImgCoord\right)}-\sImgWeights_{t}\left(\sImgCoord\right)  &\text{otherwise}
\end{cases}
,
\end{equation}
where $\psi(\sImgCoord)$ is the corresponding point index that is projected on the pixel location $\sImgCoord$ on both $\sRefMasks_{t}(\sImgCoord)$ and $\sRefWeights_{t}(\sImgCoord)$. In particular, when $\sWeight_{\psi(\sImgCoord)}<0$, we set the entity label $\sLabel_{\psi(\sImgCoord)}$ to $\sConsistImgMasks_{t}(\sImgCoord)$, and the weight $\sWeight_{\psi(\sImgCoord)}$ to $\sImgWeights_{t}(\sImgCoord)$.\par%
%
%%%%%%%%%%%%%%%%%%%%%%%%%%%%%%%%%%%%%%%%%%%%%%%%%%%%%%%%%%%%%%%%%%%%%%%%%%%%%%%%%%%%%%%%%%%
\subsubsection{Extraction}\label{sec:property_extraction}%
% Given a labeled point map and a set of keyframes, we extract 3D bounding boxes of entities, the entity visibility graph $\sCoVisGraph$, and the neighbour graph $\sProxGraph$ of the current reconstruction.\par%
%
We use the points belonging to each entity label to compute the 3D OBB $\sBBox_{i}$ of an entity $\sNode_{i}\in\sNodes$. We perform statistical outlier removal (from PCL~\cite{Rusu_ICRA2011_PCL}) to filter out points that could lead to distorted boxes. For the computation, we make use of the minimum volume estimation method~\cite{chang2011} assuming gravity alignment.\par%
The entity visibility graph $\sCoVisGraph = (\sNodes, \sKFs, \sEdges_c)$ consists of all nodes $\sNodes$ and keyframes $\sKFs$ connected by visibility edges $\sEdges_c$. A visibility edge $e_{ij}\in\sEdges_c$ exists if entity $\sNode_i \in \sNodes$ is visible in keyframe $\sKF_j \in \sKFs$. Specifically, the visibility is determined by checking if any point in node $\sNode_i$ is visible at $\sKF_j$.\par%
The neighbour graph \(\sProxGraph = (\sNodes, \sEdges_p)\) consists of nodes $\sNodes$ and its proximity edges $\sEdges_p$. A proximity edge $e_{i\to j}\in\sEdges_p\mid \sNode_i,\sNode_j\in\sNodes, i\neq j$ exists if nodes $\sNode_i, \sNode_j$ are close in space, which is determined using a bounding box collision detection method. Since the size of the OBBs is not precise, we extend their dimensions by a margin $\sThresColli$ to include additional potential neighbours.\par%
%
%%%%%%%%%%%%%%%%%%%%%%%%%%%%%%%%%%%%%%%%%%%%%%%%%%%%%%%%%%%%%%%%%%%%%%%%%%%%%%%%%%%%%%%%%%%
%%%%%%%%%%%%%%%%%%%%%%%%%%%%%%%%%%%%%%%%%%%%%%%%%%%%%%%%%%%%%%%%%%%%%%%%%%%%%%%%%%%%%%%%%%%
\subsection{Semantic Scene Graph Prediction}\label{sec:network}%
For every of the scene extractions obtained by the IEE front end, SSGP estimates 3D semantic scene graphs using message passing to jointly update initial feature representations and relationships ~\cite{Xu2017,Gay2018_visual,Wald2020_3dssg,wu2021scenegraphfusion}. In the last step, the network fuses all of them into a consistent global 3D SSG. %
The initial node features are computed with multi-view image features (\SEC{node_feature}), while the initial edge feature is computed with the relative geometric properties of its connected two nodes (\SEC{edge_feature}). Both initial features are jointly updated with a GNN along the connectivity given by the neighbour graph (\SEC{message_passing}). The updated node and edge are used to estimate their class distribution (\SEC{classification_loss}). We apply a temporal scene graph fusion procedure to combine the predictions into a global 3D SSG (\SEC{fusion}).\par
Our network architecture combines the benefits of 2D and 3D scene graph estimation methods by using 2D image features and 3D edge embedding. Image features are generally a better scene representation than 3D features, while using edge embedding in 3D allows performing relationship estimation without the constraint of the field of view. The effects of 2D and 3D features are compared in \SEC{eval_3dssg}. \par%
% Our network architecture combines the benefits from 2D and 3D scene graph methods by the use of image feature and 3D edge embedding. The primary benefit of using multi-view image features is twofold. First, it considers object observations from all given input views. Unlike VGfM~\cite{Gay2018_visual}, which only consider frames within a short time range. Second, it is computationally efficient in updating the node feature when the number of views increases. The use of edge embedding in 3D allows relationship estimation without the constraint of the field of view.\par%
%
%%%%%%%%%%%%%%%%%%%%
\subsubsection{Node Feature}\label{sec:node_feature}%
For each node $\sNode_{i}\in\sNodes$, we compute a multi-view image feature $\sNodeImgFeature_{i}$ and a geometric feature $\sNodeGeoFeature_{i}$. We use the former as the initial node feature $\sNodeFeature_{i}=\sNodeImgFeature_{i}$ and include the latter with a learnable gate in the message passing step (\SEC{message_passing}). \par%
The multi-view image feature is computed by aggregating multiple observations of $\sNode_{i}$ on images given by the entity visibility graph. For each view, an image feature is extracted with an image encoding network given the Region-of-Interest (ROI) of the node. The image features are aggregated using a mean operation to the multi-view image feature $\sNodeImgFeature_{i}$. Although there are sophisticated methods to compress multi-view image features, such as using gated averaging~\cite{Gay2018_visual} and learning a canonical representation~\cite{wei2021learning_cvr}, we empirically found that averaging all the input features~\cite{su2015multi_mvcnn} yields the best result (see supplementary material). The mean operation also allows incrementally computing the multi-view image feature with a simple moving average. The geometric feature $\sNodeGeoFeature_{i}$ is computed from the point set $\sPoints_i$ using a simple point encoder~\cite{Ruizhongtai2016_pointnet}.\par%
%
%%============================================================
\subsubsection{Edge Feature}\label{sec:edge_feature}
For each edge $\sEdge_{i\to j}\in\sEdges_{p}$, an edge feature $\sEdgeFeature_{i\to j}$ is computed using the node properties from its connected two nodes $\sNode_i$ and $\sNode_j$ by
\begin{equation}\label{eq:edge_feature}
    \sEdgeFeature_{i\to j} = \sMLPEdge \left(\left[ 
    \sCenter_j-\sCenter_i,\sBBoxSize_j-\sBBoxSize_i,\sRotationFeature_{i\to j} \right]\right),
\end{equation}
where \(\sMLPEdge \left(\cdot\right)\) is a Multilayer Perceptron (MLP), $[\cdot]$ denotes a concatenation function, and $\sRotationFeature_{i\to j}$ is a relative pose descriptor which encodes the relative angle between two entities.\par
\begin{figure}
    \centering
    \includegraphics[width=0.8\columnwidth]{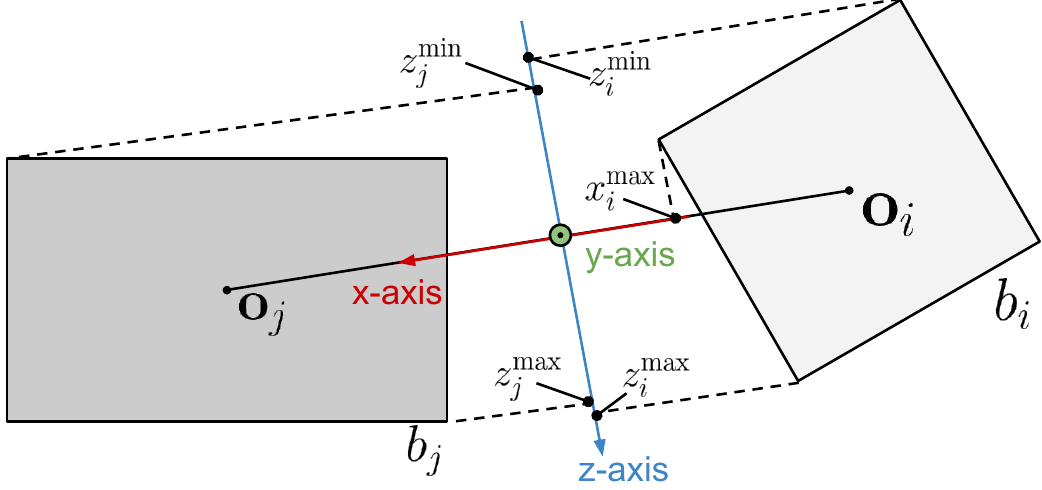}
    \caption{An illustration of our relative pose descriptor. 
    The descriptor describe the relative maximum and minimum value of given two bounding boxes on a reference frame.
    % We first find the mid point of two BBs. The x-axis is defined as the mid point to the $\sCenter_j$, the y-axis (point at readers direction) is the reverse gravity direction, and the z-axis is the cross product of x and x axes. For simplicity, we only show the projections on z-axis, and a projection on x-axis.
    }
    \label{fig:rel_pose_desc}
    \vspace{-0.3cm}
\end{figure}

%
The relative pose descriptor is designed to implicitly encode relative angles between two nodes. Using an explicit one is not optimal since OBB estimations do not return the exact pose of an object, which makes explicit pose descriptor not applicable. We instead use the relative geometry properties on a reference frame constructed by the two nodes to implicitly encode the relative pose, as illustrated in \FIG{rel_pose_desc}. First, we construct a reference frame with the origin the midpoint of the center of two nodes, the x-axis to $\sCenter_{j}$, the y-axis to the inverse of the gravity direction, and the z-axis the cross product of the x-axis and y-axis. Second, we take maximum and minimum values on each axis of the reference frame to compute the relative pose descriptor as
% 
% First, we construct a reference frame, with the origin at the mid-point of the two node centers, the x-axis points to $\sCenter_{j}$, the y-axis points to the inverse of the gravity direction, and the z-axis is the cross product of the x-axis and y-axis. Then, we take the log-absolute ratio of the maximum and minimum axes values from the projected bounding box boundaries on the reference frame as the implicit relative angle encoding:%
\begin{equation}
    R_{i\to j} = \log\left(\left|\left[\sPoint^{\text{max}}_i \oslash \sPoint^{\text{max}}_j,\sPoint^{\text{min}}_i \oslash \sPoint^{\text{min}}_j\right]\right|\right),
\end{equation}
where $\oslash$ is the Hadamard division, $\sPoint^{\text{max}}_{\square}, \sPoint^{\text{min}}_{\square} \in\mathbb{R}^{3}$ are the maximum and minimum points on the reference frame for $\square\in(i,j)$. We use an absolute logarithm ratio to improve the numerical stability. 
% \begin{equation}
%     R_{i\to j}=\log\left(\left[|\frac{x^{\text{max}}_i}{x^{\text{max}}_j}|,|\frac{x^{\text{min}}_i}{x^{\text{min}}_j}|,|\frac{y^{\text{max}}_i}{y^{\text{max}}_j}|,|\frac{y^{\text{min}}_i}{y^{\text{min}}_j}|,|\frac{z^{\text{max}}_i}{z^{\text{max}}_j}|,|\frac{z^{\text{min}}_i}{z^{\text{min}}_j}|\right]\right)
% \end{equation}
% where $\square^{\text{max}}$ and $\square^{\text{min}}$ are the maximum and minimum values of the projected bounding box boundaries on $\square$ axis of the reference frame for either node $i$ or $j$.
%
% https://docs.google.com/drawings/d/121eiDgLbtk6jfMqSr_QQpCajrBk6zKUjMZbOT9E0jHc/edit

%
%%%%=============================================================================
\subsubsection{Message Passing}\label{sec:message_passing}
Given an initial node feature $\sNodeFeature_{i}$ and an edge feature $\sEdgeFeature_{i\to j}$, we aggregate the messages from the neighbors for both nodes and edges to enlarge the receptive field and leverage the spatial understanding composition of the environment. We follow~\cite{Xu2017} by aggregating the messages with a respective GRU unit shared for all nodes and edges. Following, we explain the process taking place in each of the message-passing layers.\par%
First, we incorporate the geometric feature to each node feature using a learnable gate:
\begin{equation}
    \sNodeFeaturePlus_{i} = \sNodeFeature_{i}+\sigma \left(\mathbf{w}^{T}[\sNodeFeature_{i},\sNodeGeoFeature_{i}] \right) \sigma(\sNodeGeoFeature_{i}),
\end{equation}
where $\sNodeFeaturePlus_{i}$ is the enhanced node feature, $\sigma$ denotes a sigmoid function, and $\mathbf{w}^{T}$ are learnable parameters. The geometric feature may be unreliable, especially when the input geometry is ambiguous or unstable. Thus, we use the learnable gate to learn if the feature should be included. A node message $m_i$ and an edge message $m_{i\to j}$ are computed by
\begin{gather}
    m_i = \sMLPGCNNode 
    \left(\left[ 
        \sNodeFeaturePlus_{i}, \max_{j\in \funcNeighbors{i}}\left( \sFAN\left( \sNodeFeaturePlus_{i},\sEdgeFeature_{i\to j},\sNodeFeaturePlus_{j} \right) \right) 
    \right]\right),\\
    m_{i\to j} = \sMLPGCNEdge \left([ \sNodeFeaturePlus_{i}, \sEdgeFeature_{i\to j}, \sNodeFeaturePlus_{j} ]\right),
\end{gather}
where \(\sMLPGCNNode \left(\cdot\right)\) and \(\sMLPGCNEdge \left(\cdot\right)\) are MLPs, $\funcNeighbors{i}$ is the set of indices representing the neighbouring nodes of $i$, $\sFAN$ is the feature-wise attention network~\cite{wu2021scenegraphfusion} which weights all neighbour node feature $\sNodeFeaturePlus_{j}$ using input query \(\sNodeFeaturePlus_{i}\), and key \(\sEdgeFeature_{i\to j}\) given \(j \in \funcNeighbors{i}\).%
% An edge message $m_{i\to j}$ is computed by
% \begin{equation}
%     m_{i\to j} = \sMLPGCNEdge \left([ \sNodeFeaturePlus_{i}, \sEdgeFeature_{i\to j}, \sNodeFeaturePlus_{j} ]\right).
% \end{equation}\par%
% Both node and edge features are updated with a respective GRU unit~\cite{Xu2017}, which are shared for all nodes and edges.\par%
%
%%%%=============================================================================
\subsubsection{Class Prediction and Loss Functions}\label{sec:classification_loss}
We use the softmax function to estimate the class distribution on both nodes and edges. For multiple predicate estimations, we use the sigmoid function (with a threshold of 0.5) to estimate whether a predicate exists. The network is trained with a cross-entropy loss for classifying both entities and edges. The loss for the edge class is replaced with binary cross entropy for multiple predicates estimation~\cite{Wald2020_3dssg}.\par%
\subsubsection{Fusion}\label{sec:fusion}%
Multiple predictions on the same nodes and edges are fused to ensure temporal consistency. We use the running average approach~\cite{curless1996volumetric} to fuse predictions~\cite{wu2021scenegraphfusion}. For each entity and edge, we store the full estimated probability estimation $\sProbs^{t}$ and a weight $\sProbWeight^{t}\in\mathbb{R}_{\ge0}$ at time $t$. Given a new prediction, we update the previously stored $\sProbs^{t-1}$ and $\sProbWeight^{t-1}$ as
\begin{gather}
    \sProbs^{t} = \frac{\sProbs^{t}\cdot \sProbWeight^{t} + \sProbs^{t-1}\cdot \sProbWeight^{t-1}}{\sProbWeight^{t}+\sProbWeight^{t-1}},\\
    \sProbWeight^{t} = \min\left( \sProbWeight_{\text{max}}, \sProbWeight^{t} + \sProbWeight^{t-1} \right ),
\end{gather}
where $\sProbMaxWeight$ is the maximum weight value.\par%
\section{Evaluation}\label{sec:evaluation}
% INTRO
We evaluate our method on the task of 3D semantic scene graph estimation (\SEC{eval_3dssg}) and incremental label association (\SEC{eval_label}). In addition, we provide ablation studies on the proposed network (\SEC{eval_abla}), and a runtime analysis of our pipeline (\SEC{runtime}).\par% We will first detail the experiment setup, then report the evaluation results and ablation studies.\par%

% \paragraph{Experiment Setup.}
\subsection{Implementation Details}\label{sec:impl}
In all experiments, we use the default ORB-SLAM3~\cite{campos2021orb3} setup provided by the authors\footnote{\url{https://github.com/UZ-SLAMLab/ORB_SLAM3.git}} for our IEE front end. 
For the 2D entity detection, we use EntitySeg~\cite{qi2021entity} with a ResNet50~\cite{he2016deep} backbone pretrained on COCO~\cite{lin2014microsoft} and fine-tuned on the 3RScan~\cite{Wald2019RIO} training split. For multi-view feature extraction, we use a ResNet18~\cite{he2016deep} pretrained on ImageNet~\cite{deng2009imagenet} without fine-tuning. The point encoder is the vanilla PointNet without learned feature transformation~\cite{Ruizhongtai2016_pointnet}. 
Regarding hyperparameters, we set $\sIoUThres$ to 0.2, $\sThresColli$ to 0.5 meters, $\sProbMaxWeight$ to 100, and the number of message passing layers to 2. 

We use the ground truth pose to guide the scene reconstruction because (i) our focus lands on entity detection and scene graph estimation, and (ii) the provided image sequence from 3RScan~\cite{Wald2019RIO} has a low frame rate (10 Hz), severe image blur, and jittery motion.\par
% ================================================================================================
\subsection{3D Semantic Scene Graph estimation}\label{sec:eval_3dssg}
%%%% Triplet Recall, mRecall (Instance level)
\setlength{\tabcolsep}{4.0pt}
\begin{table}[t]
\begin{center}%
\begin{tabular}{c|l|rrr|rr}%
\toprule%
\multicolumn{1}{c}{~} & \multicolumn{1}{c}{\multirow{2}{*}{Method}} & \multicolumn{3}{c}{\emph{Recall}(\%)} & \multicolumn{2}{c}{\emph{mRecall}(\%)} \\
\multicolumn{1}{c}{~} & \multicolumn{1}{c}{~}  &  \multicolumn{1}{c}{Rel.} & Obj. & \multicolumn{1}{c}{Pred.} & \multicolumn{1}{c}{Obj.} & Pred.\\
\midrule%
\multirow{5}{*}{\rotatebox{90}{\emph{GT}}} 
        & IMP~\cite{Xu2017} & 49.8 & 70.1 & 94.3 & 53.0 & 38.1\\%IMP_full_l20_6
        & VGfM~\cite{Gay2018_visual} & 49.3 & 69.4 & 94.8 & 57.5 & 44.6\\%VGfM_full_l20_8
        & 3DSSG~\cite{Wald2020_3dssg} & 34.6 & 58.0 & 95.2 & 46.8 & 58.7\\%3DSSG_full_l20_3
        & SGFN~\cite{wu2021scenegraphfusion} & 41.8 & 63.8 & 94.3 & 57.7 & 65.5\\%SGFN_full_l20_2
        & Ours & \textbf{66.1} & \textbf{81.2} & \textbf{95.6} & \textbf{77.4} & \textbf{71.5}\\%JointSSG_full_l20_5
        \midrule
\multirow{5}{*}{\rotatebox{90}{\emph{Dense}}} 
        & IMP~\cite{Xu2017} & 25.8 & 51.8 & 90.4 & 30.0 & 23.0\\%IMP_INSEG_l20_3
        & VGfM~\cite{Gay2018_visual} & 28.3 & 53.3 & \textbf{90.7} & 31.6 & 24.4\\%VGfM_inseg_l20_1
        & 3DSSG~\cite{Wald2020_3dssg} & 17.5 & 41.4 & 88.2 & 31.9 & 26.6\\%3DSSG_INSEG_l20_3
        & SGFN~\cite{wu2021scenegraphfusion}  & 31.4 & 56.7 & 89.6 & 38.3 & 30.5\\%SGFN_inseg_l20_0
        & Ours & \textbf{34.1} & \textbf{58.1} & 89.9 & \textbf{43.0} & \textbf{33.3}\\%Joint_inseg_l20_1
        \midrule
\multirow{6}{*}{\rotatebox{90}{\emph{Sparse}}} 
        & IMP~\cite{Xu2017} & 7.9 & 27.5 & 90.7 & 20.6 & 14.0\\%IMP_orbslam_l20_1
        & VGfM~\cite{Gay2018_visual} & 8.2 & 26.9 & \textbf{90.8} & 17.6 & 15.4\\%VGfM_orbslam_l20_1
        & 3DSSG~\cite{Wald2020_3dssg} & 0.9 & 9.7 & 87.9 & 5.9 & 15.1\\%3DSSG_3rscan_orbslam_l20_1
        & SGFN~\cite{wu2021scenegraphfusion} & 1.7 & 12.6 & 88.9 & 8.3 & 14.4\\%SGFN_3rscan_orbslam_l20_1
        & Ours & 9.9 & 29.5 & 90.4 & 23.5 & \textbf{16.5}\\%JointSSG_orbslam_l20_11_4
        & Ours (i) & \textbf{10.7} & \textbf{30.2} & 90.4 & \textbf{24.5} & 15.9\\%JointSSG_orbslam_l20_11_4
\bottomrule
\end{tabular}
\end{center}
\caption{We compare our method with four baseline methods on the task of scene graph prediction on 3RScan~\cite{Wald2019RIO} dataset with 20 objects and 8 predicate classes. The results from Ours are obtained by using our network to obtain predictions, while Ours (i) contains the results from using the incremental pipeline.}
\label{tbl:3dssg_full_eval_l20} 
\end{table}%%
%%%% Change to instance level eval.
%\begin{wraptable}{r}{7.5cm}
\begin{table}[t]
%\vspace{-0.8cm}
% \begin{table}[]
    \centering
% \resizebox{\linewidth}{!}{
    \begin{tabular}{l|ccc|cc}
    \toprule
      \multicolumn{1}{c}{\multirow{2}{*}{Method}}     & \multicolumn{3}{c}{\emph{Recall}(\%)}       & \multicolumn{2}{c}{\emph{mRecall}(\%)} \\
    \multicolumn{1}{c}{~}  &  \multicolumn{1}{c}{Rel.} & Obj. & \multicolumn{1}{c}{Pred.} & \multicolumn{1}{c}{Obj.} & Pred.\\
    \midrule
        IMP~\cite{Xu2017} & 44.5 & 35.9 & 9.0 & 18.7 & 4.9\\%IMP_FULL_l160_2_3
        VGfM~\cite{Gay2018_visual} & 44.5 & 37.9 & 14.7 & 17.9 & 6.5\\%VGfM_FULL_l160_3_1
        3DSSG~\cite{Wald2020_3dssg} & 46.8 & 29.6 & \textbf{68.8} & 11.7 & \textbf{25.5}\\%3DSSG_full_l160_1
        SGFN~\cite{wu2021scenegraphfusion} & 45.2 & 29.4 & 42.8 & 11.8 & 13.5\\%SGFN_full_l160_4
        Ours & \textbf{52.7} & \textbf{56.7} & 50.4 & \textbf{27.2} & 23.9\\%JointSSG_full_l160_0
        \bottomrule
    \end{tabular}
% }
\caption{Evaluation on scene graph prediction with 160 object and 26 predicate classes using ground truth segmentation and fully connected neighbor graph.
}
\label{tbl:3dssg_full_l160}
\vspace{-0.3cm}
\end{table}
%\end{wraptable}%
For the input types, we compare all methods with the input of ground truth segmentation~\cite{Wald2020_3dssg} (\emph{GT}), geometric segmentation~\cite{tateno2017large} (\emph{Dense}) and sparse segmentation (\emph{Sparse}). For the baseline methods, we compare ours with two 2D methods (IMP~\cite{Xu2017}, and VGfM~\cite{Gay2018_visual}), and two 3D methods (3DSSG~\cite{Wald2020_3dssg} and SGFN~\cite{wu2021scenegraphfusion}).\par%
\noindent\textbf{Baseline Methods.}
We will briefly discuss baseline methods here. Check supplementary for further details. %
IMP~\cite{Xu2017} computes a node feature using the image feature cropped from the ROI of the node in an image and computes an edge feature using the union of two ROIs from its connected nodes. Both features are jointly updated with prime-dual message passing and learnable message pooling. %
VGfM~\cite{Gay2018_visual} extends IMP by adding geometric features and temporal message passing to handle sequential estimation. %
3DSSG~\cite{Wald2020_3dssg} extends the ROI concept in IMP~\cite{Xu2017} in 3D by replacing ROIs to 3D bounding boxes. The node and edge features are computed with PointNet~\cite{Ruizhongtai2016_pointnet}. Both features are jointly updated with a graph neural network with average message pooling. %
SGFN~\cite{wu2021scenegraphfusion} improves 3DSSG by replacing the initial edge descriptor with the relative geometry properties between two nodes and introducing an attention method to handle dynamic message aggregation, which enables incremental scene graph estimation.\par%
\noindent\textbf{Implementation.}
For all methods, we follow their implementation details and train on 3RScan dataset~\cite{Wald2019RIO} from scratch until converge, using a custom training and test split since the scenes in the original test do not have ground truth scene graphs provided. For IMP~\cite{Xu2017}, since it is a single image prediction method, we adopt the voting mechanism as in~\cite{kim20193} to average the prediction over multiple frames. Since ours and other 2D baseline methods rely on image input, we generate a set of keyframes by sampling all input frames using their poses for the \emph{GT} and \emph{Dense} inputs (check supplementary material for further details). To ensure diversity in the viewpoint, we filter out a frame if its pose is too similar to any selected frames with the threshold values of 5 degrees in rotation and 0.3 meters in translation.\par%
\noindent\textbf{Evaluation Metric.}
We report the overall recall (\emph{Recall}) as used in many scene graph work~\cite{lu2016visual,Xu2017,Yang2018,Wald2020_3dssg,wu2021scenegraphfusion} but with the strictest top-k metric with $k=1$ as in \cite{wald2022learning}. In addition, we report the mean recall (\emph{mRecall}) which better indicates model performance when the input dataset has a severe data imbalance issue (see supplementary material for the class distribution). Moreover, since different segmentation methods may result in different number of segments, we map all predictions on estimated segmentation back to ground truth. This allows us to compare the reported numbers across different segmentation methods. We report the \emph{Recall} of relationship triplet estimation (Rel.), object class estimation (Obj.) and predicate estimation (Pred.), and the \emph{mRecall} of object class estimation and predicate estimation.\par%
\begin{figure}[t]
    \centering
    %\resizebox{\columnwidth}{!}{
    \includegraphics[width=\columnwidth]{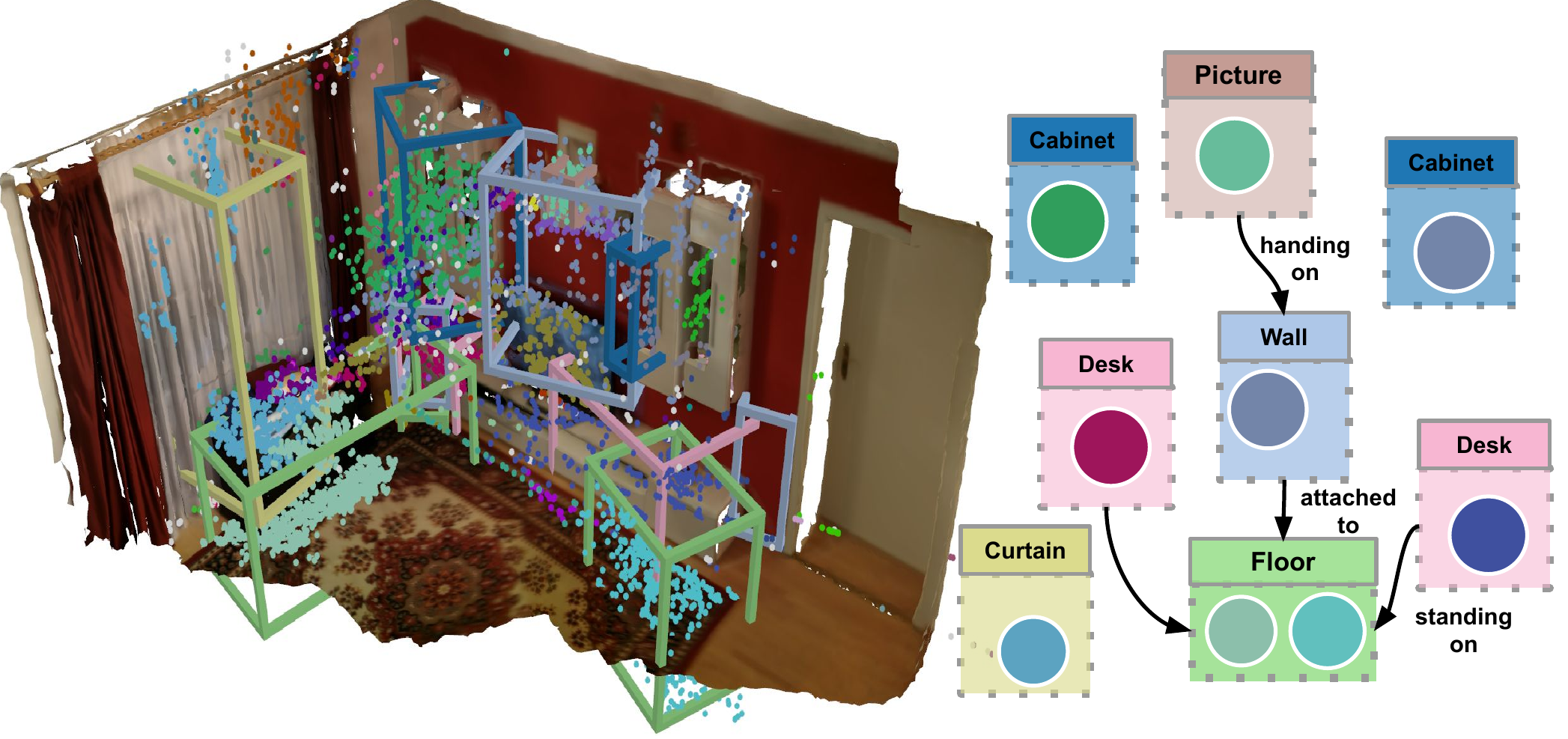}
    %}
    \caption{Qualitative evaluation of our scene graph prediction framework. Each 3D bounding box represents a detected entity on the left, and the color is the predicted label. On the right side, we visualize the estimated scene graph on this scene. We only select representative entities on the scene graph visualization for visualization purposes.}
    \label{fig:qualitative}
\vspace{-0.5cm}
\end{figure}%
%
% We compare our method to two 2D methods \ie~IMP~\cite{Xu2017} VGfM~\cite{Gay2018_visual}, and two 3D methods, \ie~Wald et al~\cite{Wald2020_3dssg} and SGFN~\cite{wu2021scenegraphfusion}.
%
%
\noindent\textbf{Results.}
Following the evaluation scheme in \cite{wu2021scenegraphfusion} and in \cite{Wald2020_3dssg}, we report two evaluations in \TBL{3dssg_full_eval_l20} and \TBL{3dssg_full_l160}, respectively. The former one maps the node classes to 20 NYUv2 labels~\cite{Silberman2012} to suppress the severe class imbalance in the data as discussed in~\cite{wald2022learning} and estimates a single predicate out of seven support relationship types plus the ``same part'' relationship to handle over-segmentation. The latter uses 160 node and 26 edge classes with multiple predicate estimation.\par% 
In \TBL{3dssg_full_eval_l20}, overall, it can be seen that all image-based methods (IMP, VGfM, Ours) outperforms points-based methods (3DSSG, SGFN) in almost all object prediction metrics, while the methods based on 3D edge descriptor (3DSSG, SGFN, Ours) tend to have better predicate estimation. This suggests that the 2D representations from images are more representative than 3D, and 3D edge descriptors are more suitable for estimating support types of predicates. 
By comparing IMP~\cite{Xu2017} and VGfM~\cite{Gay2018_visual}, it can be seen that the effect of the geometric feature and the temporal message passing is mainly reflected in the \emph{mRecall} metrics. However, it deteriorates the performance when the input is sparse. A possible reason is that the geometric feature is relatively unstable, which decreases the network performance. It is also reflected in the two 3D methods, \ie 3DSSG and SGFN, where they failed to perform in classifying objects with sparse segmentation while giving a similar performance in the predicate classification.
Among all methods, our method outperforms all baselines among all input types and all metrics, apart from the predicate estimation, which has a slightly worse result on some input types. In addition, we report ours using the proposed incremental estimation pipeline, denoted as Ours(i). The incremental estimation process improves slightly in object estimation. The same behavior is also reported in \cite{wu2021scenegraphfusion}. We show a qualitative result using our full pipeline in \FIG{qualitative}.\par%
In \TBL{3dssg_full_l160}, our method outperforms all other methods in the relationship and object estimation in \emph{Recall} and object estimation in \emph{mRecall}. 3DSSG~\cite{Wald2020_3dssg} has the best results in predicate estimations. This suggests that union 3D bounding boxes are more suitable when estimating multiple predicates.\par%
%
% The full evaluation is reported in  \TBL{3dssg_full_eval_l20} and \TBL{3dssg_full_l160}.%
%%%%%%%%%%%%%%%%%%%%%%%%%%%%%
% with GT segmentation
%%%%%%%%%%%%%%%%%%%%%%%%%%%%%
% ================================================================================================
\subsection{Incremental Label Association}\label{sec:eval_label}
We evaluate our label association method in the task of incremental entity segmentation, which aims to estimate accurate class-agnostic segmentation given sequential sensor input, with two baseline methods, \ie InSeg~\cite{Tateno2015} and PanopticFusion~\cite{Gaku2019_panopticfusion}.\par%
\noindent\textbf{Baseline Methods.}
Both baseline methods use reference map approaches as mentioned in \SEC{entity_fusion}. InSeg~\cite{Tateno2015} considers only the overlapping ratio between labels on an estimated mask and a reference mask. PanopticFusion uses IoU~\cite{Gaku2019_panopticfusion} as the evaluation method and limits one reference label can only be assigned to one query label.\par%
\noindent\textbf{Implementation.}
For all methods, we use our IEE pipeline with different label association methods on all training scenes in the 3RScan dataset~\cite{Wald2019RIO}.\par%
\noindent\textbf{Evaluation Metric.}
We use the Average Overlap Score (AOS) as the evaluation metric~\cite{tateno2017large}. It measures the ratio of the dominant segment of its corresponding ground truth instance. %
We use the nearest neighbour search to find the ground truth instance label of a reconstructed point. Since our method reconstructs a sparse point map, using the ground truth points as the denominator does not reflect the performance. We instead calculate the score over the sum of all estimated points within the ground truth instance as:%
\begin{equation}
    \text{AOS} = \sum_i \frac{\max_{j} \text{Overlap}(\mathcal{S}_i,\sPoints_j) }{\funcCardi{\mathcal{S}_i}},
\end{equation}%
where $\mathcal{S}_i$ is the set of all estimated points with ground truth instance label $i$.\par%

\noindent\textbf{Results.}
The evaluation result is reported in \TBL{comp_iee}. Our approach achieves the highest AOS, which is 1 \% higher than InSeg~\cite{Tateno2015} and 3.7\% higher than PanopticFusion~\cite{Gaku2019_panopticfusion}. The use of a confidence-based approach handles better the label consistency and thus improves the final AOS score. We provide an example of how our method improves temporal consistency under non-uniform distributed points in the supplementary material.\par%
\begin{table}[t]
    \centering
    \begin{tabular}{l|c}
         Method &  \text{AOS} (\%) \\
         \hline
         InSeg~\cite{Tateno2015} &  38.6\\
         PanopticFusion~\cite{Gaku2019_panopticfusion} & 35.9\\
         Ours & \textbf{39.6}\\
    \end{tabular}
    \caption{Evaluation of different label association methods in the task of incremental entity estimation in 3RScan dataset~\cite{Wald2019RIO}.}
    \label{tbl:comp_iee}
    \vspace{-0.5cm}
\end{table}%
\subsection{Ablation Study}\label{sec:eval_abla}
We ablate our network with two components, \ie geometric descriptor $\sNodeGeoFeature_i$ and relative pose descriptor $\sRotationFeature_{i\to j}$. The experiment setup is the same as in \cite{wu2021scenegraphfusion}, which makes the ablation comparable to \TBL{3dssg_full_eval_l20}. The result is reported in \TBL{abla}. More ablation studies are in the supplementary material.\par%
Our vanilla network without $\sNodeGeoFeature_i$ and $\sRotationFeature_{i\to j}$ outperform baselines in most of the metrics. With $\sNodeGeoFeature_i$, there is a consistent improvement on all metrics except the Pred. in \emph{mRecall}. Compared to VGfM~\cite{Gay2018_visual} in \TBL{3dssg_full_eval_l20}, VGfM~\cite{Gay2018_visual} fails to improve the performance of IMP~\cite{Xu2017} with sparse input. Our gated geometric feature aggregation improves its baseline with sparse input, bringing a more consistent performance gain than VGfM~\cite{Gay2018_visual}. The $\sRotationFeature_{i\to j}$ improves the \emph{mRecall} performance with \emph{GT} and \emph{Dense} inputs but decreases the model \emph{Recall} performance. This behavior suggests that $\sRotationFeature_{i\to j}$ helps handle class imbalance issues. The combination of both components achieves the best \emph{Recall}. However, the model tends to focus on dominant classes, resulting in slightly worse performance in \emph{mRecall}.\par%
\begin{table}[t]
    \centering
    \begin{tabular}{c|cc|rrr|rr}
    \toprule
    \multicolumn{1}{c}{~} & \multicolumn{2}{c}{Method} & \multicolumn{3}{c}{\emph{Recall}(\%)} & \multicolumn{2}{c}{\emph{mRecall}}\\
    \multicolumn{1}{c}{~} & $\sNodeGeoFeature_i$ & \multicolumn{1}{c}{$\sRotationFeature_{i\to j}$} & Rel. & Obj. & Pred. & Obj. & Pred. \\
    \midrule
    \multirow{4}{*}{\rotatebox{90}{\emph{GT}}} 
    & & & 61.9 & 76.4 & 95.6 & 74.3 & 69.2\\%JointSSG_full_l20_5_1
    & \checkmark & & 62.9 & 77.9 & \textbf{95.9} & 74.2 & 64.3\\%JointSSG_full_l20_5_2
    & & \checkmark & 60.4 & 76.3 & 95.0 & 75.3 & \textbf{73.2}\\%JointSSG_full_l20_5_3 
    & \checkmark & \checkmark & \textbf{66.1} & \textbf{81.2} & 95.6 & \textbf{77.4} & 71.5\\%JointSSG_full_l20_5
    \midrule
    \multirow{4}{*}{\rotatebox{90}{\emph{Dense}}} 
    & & & 30.2 & 54.0 & 88.5 & 44.9 & 33.2\\%JointSSG_inseg_l20_4
    & \checkmark & & 33.9 & 56.4 & 89.7 & 45.7 & 33.8\\% JointSSG_inseg_l20_2
    & & \checkmark & 28.7 & 52.7 & 88.2 & \textbf{47.5} & \textbf{34.3}\\%JointSSG_inseg_l20_3
    & \checkmark & \checkmark & \textbf{34.1} & \textbf{58.1} & \textbf{89.9} & 43.0 & 33.3\\% JointSSG_inseg_l20_1
    \midrule
    \multirow{4}{*}{\rotatebox{90}{\emph{Sparse}}} 
    & & & 9.6 & 28.6 & 90.0 & 25.6 & 17.7\\%JointSSG_orbslam_l20_11_6
    & \checkmark & & 9.8 & 28.5 & 90.0 & \textbf{25.7} & \textbf{18.1}\\%JointSSG_orbslam_l20_11_7
    & & \checkmark & 9.6 & 28.1 & 90.2 & 23.3 & 16.9\\%JointSSG_orbslam_l20_11_8
    & \checkmark & \checkmark & \textbf{9.9} & \textbf{29.5} & \textbf{90.4} & 23.5 & 16.5\\%JointSSG_orbslam_l20_11_4
    \bottomrule
    \end{tabular}
    \caption{Ablation study on the proposed network. We ablate the proposed gated geometric feature ($\sNodeGeoFeature_i$) and the relative pose descriptor ($\sRotationFeature_{i\to j}$) using the same experiment setup as in \TBL{3dssg_full_eval_l20}.}
    \label{tbl:abla}
\end{table}
\subsection{Runtime}\label{sec:runtime}%
We report the runtime of our system on 3RScan~\cite{Wald2019RIO} sequence 4acaebcc-6c10-2a2a-858b-29c7e4fb410d in \TBL{runtime}. The analysis is done with a machine equipped with an Intel Core i7-8700 3.2GHz CPU with 12 threads and a NVidia GeForce RTX 2080ti GPU. 
\begin{table}[t]
\begin{center}
\resizebox{\columnwidth}{!}{
\begin{tabular}{l|c|ccc}
    \toprule
    & \emph{Front end} & \multicolumn{3}{c}{\emph{Back end}} \\
     & Sparse Mapping & \multicolumn{1}{c}{2D Entity Est.} & \multicolumn{1}{c}{Label Fusion} & Scene Graph Est. \\
    \midrule
    Mean [ms] & 14.7 & 124.6 & 14.2 & 52.5 \\
    \bottomrule
\end{tabular}
}
\caption{Runtime [ms] of the different components of our method.}
\label{tbl:runtime}
\end{center}
\vspace{-1cm}
\end{table}%
\section{Conclusion}
We present a novel method that estimates 3D scene graphs from RGB images incrementally. Our method runs in real-time and does not rely on depth inputs, which could benefit other tasks, such as robotics and AR, that have hardware limitations and real-time demand. The experiment results indicate that our method outperforms others in three different input types. The provided ablation study demonstrates the effectiveness of our design. Our vanilla network, without any geometric input and relative pose descriptor, still outperforms other baselines. Our method provides a novel architecture for estimating scene graphs with only RGB input. The multiview feature is proven to be more powerful than existing 3D methods. Our method can be further improved in many directions. In particular, using semi-direct SLAM methods such as SVO~\cite{forster2016svo} might improve the handling of untextured regions where feature-based methods often fail. In addition, the multiview image encoder could be replaced with a more powerful encoder to improve the scene graph estimation with a computational penalty.
%\newpage
%\input{sections/6_supplementary.tex}
%%%%%%%%% REFERENCES
{\small
\bibliographystyle{ieee_fullname}
\bibliography{egbib}
}

\end{document}

\title{Supplementary Material: Incremental 3D Semantic Scene Graph Prediction from RGB Sequences}
\author{Shun-Cheng Wu\textsuperscript{1}
\quad
Keisuke Tateno\textsuperscript{2}
\quad
Nassir Navab\textsuperscript{1}
\quad
Federico Tombari\textsuperscript{1,2}\\
\textsuperscript{1}Technische Universit\"{a}t M\"{u}nchen \hspace{1.5cm} \textsuperscript{2}Google\\[0.1cm]
%Boltzmannstraße 3, 85748, Germany. Brandschenkestrasse 110, 8002 Zürich\\
% {\small shuncheng.wu, johanna.wald, nassir.navab@tum.de, ktateno, tombari@google.com}}
% {\small shuncheng.wu@tum.de}
% {\small \href{https://shunchengwu.github.io/SceneGraphFusion}{shunchengwu.github.io/SceneGraphFusion}}
}
\maketitle

% For Supplementary
% \begin{todolist}
%     \item Add additional ablation study
%     \begin{todolist}
%         \item whether to use the gate in the geometric feature or to use different ways of selecting keyframes.
%     \end{todolist}
%     \item The effect of point encoder on nodes and the relative pose on edges.
%     \item implementation details
%     \item experimental design choices
%     \item How the proposed incremental method is used in a non-incremental fashion
%     \begin{todolist}
%         \item adaptations to baselines
%         \item The choice of non-learned relative pose descriptor
%         \item Network component dimensions
%         \item Keyframe selection strategy. 
%     \end{todolist}
%     \item link supplementary to the main paper.
%     \item The choice of non-learned relative pose descriptor
%     \begin{todolist}
%         \item add comparison of using fully-learned feature against ours.
%     \end{todolist}
% \end{todolist}
\setcounter{table}{5}
\setcounter{figure}{4}
\setcounter{section}{5}
% \newpage
\section{Method Comparison and Implementation}
We provide an overview of all the methods evaluated in Sec.~4.2, including our adaptation to enable comparisons among them in our experiments. As a brief recall, we compare our method to two image-based methods, \ie IMP~\cite{Xu2017} and VGfM~\cite{Gay2018_visual}, and two point-based methods, 3DSSG~\cite{Wald2020_3dssg} and SGFN~\cite{wu2021scenegraphfusion}. All methods shared a similar scene graph generation pipeline: 
\begin{enumerate}
    \item using a node and an edge encode to compute an initial node and edge embeddings.
    \item using message passing to calculate messages for updating  node and edge features.
    \item updating node and edge feature with the messages from step 2.
    \item integrating class prediction over time. 
\end{enumerate}
% 1) using a node and an edge encode to compute an initial node and edge embeddings. 2) using message passing to calculate messages for updating  node and edge features. 3) updating node and edge feature with the messages from step 2. 4) integrating class prediction over time. 
For all methods, we closely follow the original implementation from the respective papers. We refer interested readers to check out their papers for all detail. Here, we describe our adaptation to enable fair comparison in our experiments. We report the comparison of all five methods in \cref{tab:method_comp}. 

\paragraph{Object Detection.} IMP and VGfM rely on a regional proposal network to detect objects. We replace it with our entity detection methods described in Sec.~3.1.2. 

\paragraph{Geometric Features in VGfM.} For VGfM, they extract geometric features from ellipsoids. In our implementation, we replace the use of ellipsoids with oriented bounding boxes, which can provide equivalent information as needed by VGfM. 

\paragraph{Fusion for IMP.} For IMP, as mentioned in the main paper, we added the voting mechanism in \cite{kim20193} to fuse multiple predictions. For handling the incremental nature, our experiment in Table 1 follows the setup in \cite{wu2021scenegraphfusion}. We do a single global estimation of all methods and provide incremental estimations of our methods. Hence, no modification is needed for all the baseline methods.

\paragraph{Entity visibility graph and the neighbour graph for the \emph{GT} setup}
Unlike in \emph{Dense} and \emph{Sparse}, where we run an incremental estimation system to obtain $\sCoVisGraph$ and $\sProxGraph$, the \emph{GT} requires an additional procedure to obtain the entity visibility graph $\sCoVisGraph = (\sNodes, \sKFs, \sEdges_c)$ and the neighbour graph $\sProxGraph = (\sNodes, \sEdges_p)$. The entities $\sNodes$ are directly inherited from the ground truth annotation from the 3RScan dataset~\cite{Wald2019RIO}. The proximity edges $\sEdges_p$ is estimated with the same strategy as described in Sec. 3.1.4, using the ground truth bounding boxes of entities. Here we detail how we estimate $\sKFs$ and $\sEdges_c$ for the \emph{GT} setup. For $\sKFs$ and $\sEdges_c$, we follow a similar approach as in VGfM~\cite{Gay2018_visual}, where we first find all relevant frames across all entities and then select keyframes with our keyframe selection strategy (\cref{sec:keyframe_selection}). We use the ground truth instance mask from \cite{Wald2019RIO} to check if an entity appears in an image. However, an entity may be heavily occluded and cannot provide reasonable image features. To avoid this, we estimate the occupancy of an entity in an image as the ratio of the number of relevant pixels over all pixels within the bounding box of the entity. Furthermore, to prevent an entity is not aligned to the image coordinate, which will cause the occupancy value to be very low even if it is not occluded, we downscale each input mask by a factor of 8 with a maximum relevant selection, \ie a downscaled pixel is considered relevant if one among the eight pixels in the original image is relevant. This gives the visibility of each entity on each input frame. We then apply the keyframe selection strategy to prevent duplicate views and to ensure good view coverage.

\begin{table*}[ht]
    \centering
    \resizebox{\textwidth}{!}{
    \begin{tabular}{l|ccccc}
         \toprule
Method & Node Type & Edge Type & Message Passing Type & Message Update Method & Fusion \\
\midrule
IMP & Image ROI & Image ROI Union & Prime-Dual & GRU & \textcolor{cyan}{Voting}~\cite{kim20193} \\
VGfM & Image ROI & Image ROI Union & Prime-Dual + \textcolor{cyan}{Geo. description} & GRU & Temporal Gate\\
3DSSG & Points & Points Union & Triplet & concatenation & N/A\\
SGFN & Points & geometric description & FAN & concatenation & running mean\\
Ours & MV Image ROIs & geometric description & FAN + Gated Points & GRU & running mean\\ 
\bottomrule
    \end{tabular}
    }
    \caption{A summary of the modules used for different methods. The text colored in \textcolor{cyan}{cyan} involves our modification to make all methods comparable.}
    \label{tab:method_comp}
\end{table*}

\section{Keyframe selection strategy.}\label{sec:keyframe_selection}
Selecting keyframes is crucial when the estimation quality is solely based on multiview images. Having diverse view coverage of objects usually results in better feature representation of objects~\cite{yang2019learning,ho2020exploit,wei2021learning_cvr}. Unlike the keyframes selection in ORBSLM [3], which focuses more on the pose estimation quality than the view coverage, we select keyframes mainly based on the pose difference and, in addition, the quality of detected objects. A frame is selected as a keyframe only 1) it has at least one valid object detected and 2) its pose is dissimilar to other existing keyframes. We measure the validity of bounding boxes by checking if their minimum width and height are larger than 200 pixels, and the pose difference threshold is set to 5 degrees in rotation and 0.3 meters in translation. This keyframe selection method is used for all input, \ie \emph{GT}, \emph{Dense}, and \emph{Sparse}, cases in our experiments.

\section{Data Distribution}
\begin{figure*}
    \centering
    \includegraphics[width=\textwidth]{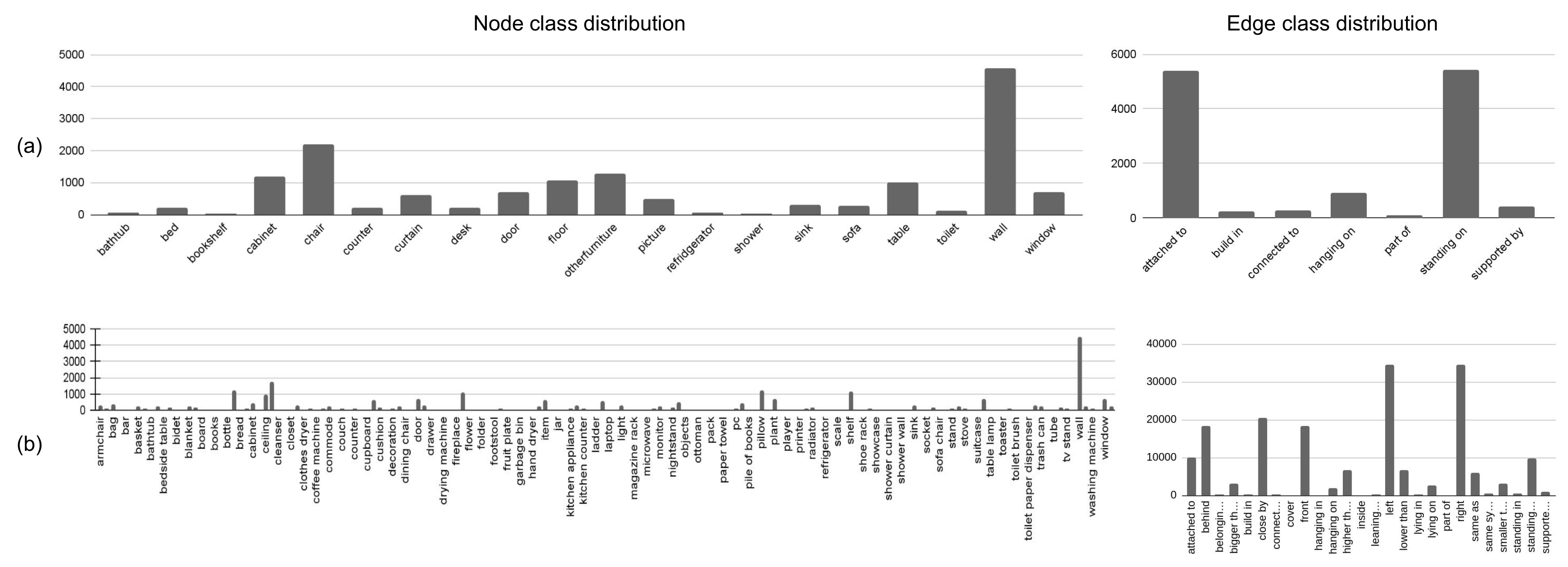}
    \caption{We provide the class distribution for two scene graph experiments in Tbl. 1 and Tbl. 2. The row (a) is the distribution for the experiment setup in \cite{wu2021scenegraphfusion} and in Tbl. 1. (b) is the distribution for the experiment setup in \cite{Wald2020_3dssg} and in Tbl. 2.}
    \label{fig:class_distribution_3rscan}
\end{figure*}

We provide the class distribution on objects and predicates in \FIG{class_distribution_3rscan}. It can be seen that the setup in \cite{Wald2020_3dssg} has severe long-tail data distribution. After mapping to 20 NYUv2 labels~\cite{Silberman2012}, the distribution is relatively well distributed but still unbalanced. The unbalanced distribution indicates that the \emph{mRecall} metric reflects better the model performance.\par%
%https://docs.google.com/drawings/d/1e30XEu4Y8Dr2hBm7i6DVtUEGMAqtgMV26_a0M-ATcxU/edit
\section{Multi-view Feature Encoding}
As mentioned in Sec. 3.2.1, the multi-view feature of a node is computed by aggregating image features of the node from multiple images. This is essentially the task of 3D shape recognition with arbitrary views. We compare the use of simple mean aggregation, as in MVCNN~\cite{su2015multi_mvcnn}, with the state-of-the-art method, CVR~\cite{wei2021learning_cvr} on ScanNet~\cite{Dai2017_scannet} dataset. Since the main comparison is on the multi-view feature aggregation, we use the same backbone, ResNet18~\cite{he2016deep}, for both methods. All networks are trained from scratch using the splits from ScanNet, following the same training approach described in the respective papers. We report the mean intersection-over-union (\emph{mIoU}), precision (\emph{mPrec}), and recall (\emph{mRecall}).\par
\begin{figure}[t]
    \centering
    %\resizebox{\columnwidth}{!}{%textwidth
    \includegraphics[width=\columnwidth]{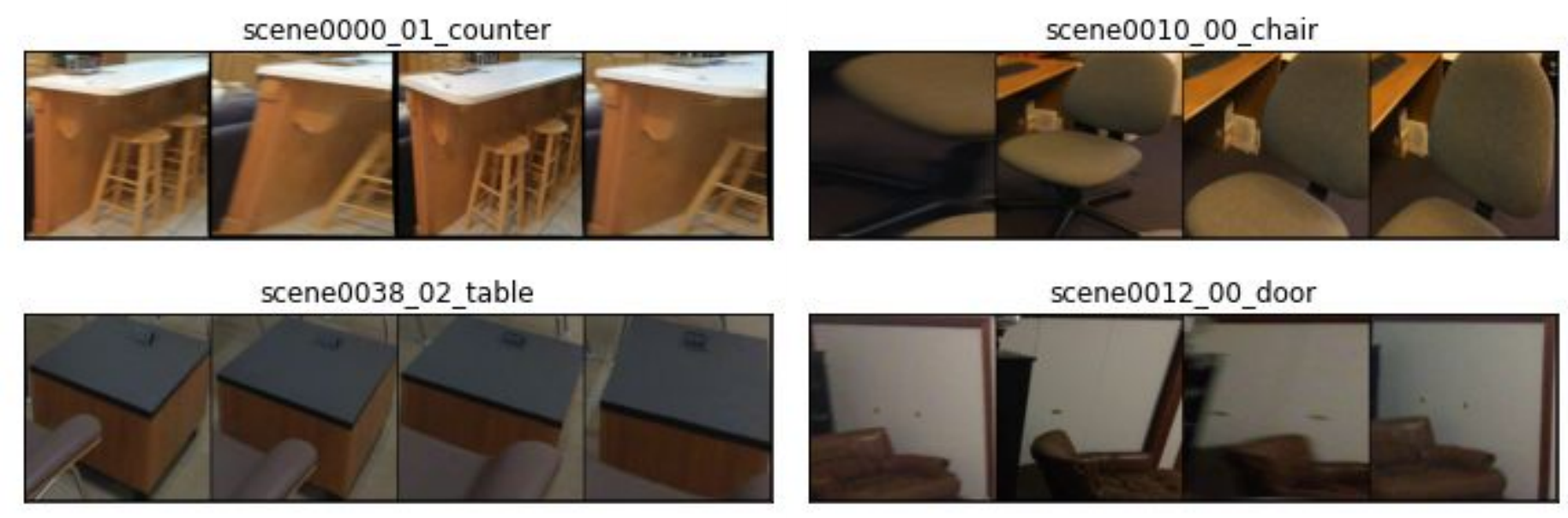}
    %}
    \caption{
    We provide examples of multi-view images of the same object in the ScanNet dataset. For display purposes, we only select four views of the same object. It can be seen that the multi-view observation of an object in a standard indoor dataset includes some non-target objects in the field of view. Those objects are considered noise which affects the multi-view image encoding.}
    \label{fig:example_mv_scannet}
\vspace{-0.3cm}
\end{figure}

%
The result is shown on \TBL{comp_mvcnn}. It can be seen that the use of simple mean operation outperforms canonical transformation in CVR~\cite{wei2021learning_cvr}. Although the number reported in CVR~\cite{wei2021learning_cvr}, it outperforms MVCNN in ModelNet40~\cite{wu20153d}, ScanObjectNN~\cite{uy2019revisiting} and RGBD~\cite{lai2011large}. We investigated the difference between the three datasets mentioned above and ScanNet and found that the images from ScanNet~\cite{Dai2017_scannet} contain more background objects while the others have non-cluttered backgrounds. We demonstrate some example images in \FIG{example_mv_scannet}. The performance inconsistency may indicate that using the mean operator makes the model less sensitive to background things.\par
\begin{table}[t]
    \centering
    \begin{tabular}{l|ccc}
    \toprule
         &  \emph{mIoU}(\%) & \emph{mPrec}(\%) & \emph{mRecall}(\%)\\
    \midrule
        CVR~\cite{wei2021learning_cvr} & 32.3 & 45.2 & 54.4\\
        MVCNN~\cite{su2015multi_mvcnn} & \textbf{39.2} & \textbf{50.5} & \textbf{61.4}\\
    \bottomrule    
    \end{tabular}
    \caption{Object classification result on ScanNet dataset. A simple averaging of overall image features (MVCNN) outperforms the sophisticated multi-view image encoding method (CVR).}
    \label{tbl:comp_mvcnn}
% \vspace{-1.5cm}
\end{table}%
\section{Comparing confidence and IoU-based methods}
\begin{figure*}[t]
    \centering
    %\resizebox{\columnwidth}{!}{
    \includegraphics[width=\textwidth]{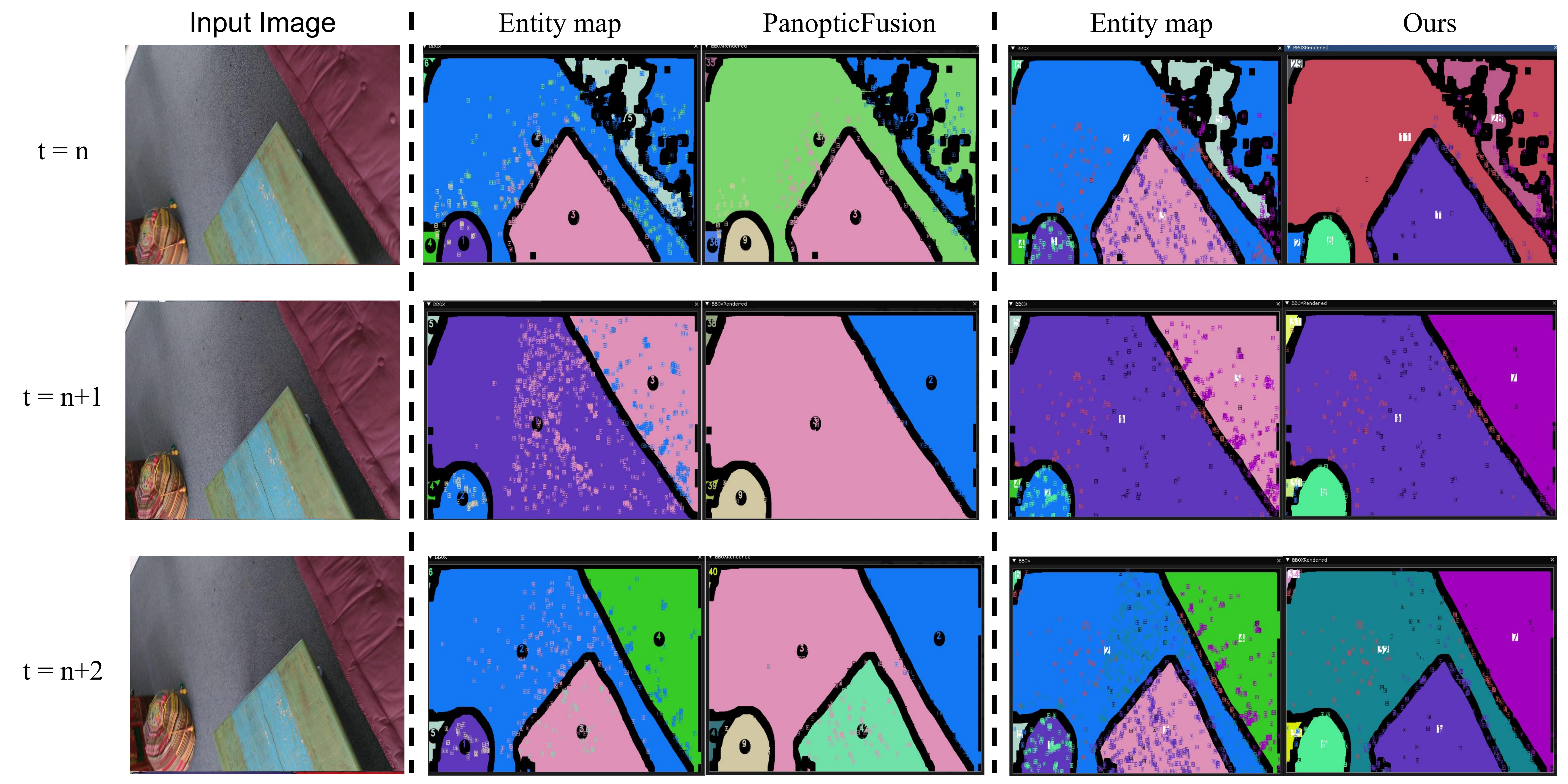}
    %}
    \caption{An example of comparing the use of IoU and our label association approach with the sparse input points. Our method handles non-uniformly distributed data better than the IoU-based method.}
    \label{fig:labelfusioncomparison}
% \vspace{-0.5cm}
\end{figure*}

We provide an example of the difference between using the maximum IoU and our maximum mean confidence to find the most probable correspondence with a sparse point map. In figure~\ref{fig:labelfusioncomparison}, given three consecutive frames at $t=n$, $t=n+1$ and $t=n+2$ for the label association and fusion, their entity maps and the association result are shown on the second and third columns (separated by white space). The second column is the label fusion result with IoU~\cite{Gaku2019_panopticfusion}, and the third column is ours. When using IoU (second column), the table label at $t=n$ is assigned to the floor at $t=n+2$. This is due to the map points created at $t=n+1$ on the floor (carpet) having larger $IoU$ than the table. With our approach (third column), the table label at $t=n$ remains at the table at $t=n+2$. This shows that our method provides more consistent label association. Note that the label colors are different between IoU and ours since the segment color are randomly in each run.\par%

\section{Ablation Study}
We provide two additional ablation studies: (1) the use of edge descriptor (\cref{tbl:abla_edge_feature}) and (2) the effect of the sigmoid gate on the geometric feature (\cref{tbl:abla_geo_gate}). In \cref{tbl:abla_edge_feature}, it can be seen that the use of our relative pose descriptor $\sRotationFeature_{i\to j}$ leads to better overall performance in \emph{GT} and \emph{Sparse} settings while having slightly worse performance in \emph{Dense} setting. In \cref{tbl:abla_geo_gate}, using the gated geometric feature consistently achieves better performance in all three setups. 

\subsection{Comparison in edge descriptor}
compare non-learned and learned descriptors.

\begin{table}[t]
    \centering
    \begin{tabular}{c|c|rrr|rr}
    \toprule
    \multicolumn{1}{c}{~} & \multicolumn{1}{c}{~} & \multicolumn{3}{c}{\emph{Recall}(\%)} & \multicolumn{2}{c}{\emph{mRecall}}\\
    \multicolumn{1}{c}{~} & \multicolumn{1}{c}{Edge Type} & Rel. & Obj. & Pred. & Obj. & Pred. \\
    \midrule
    % \multirow{2}{*}{\rotatebox{90}{\emph{GT}}} 
    \multirow{2}{*}{\emph{GT}}
    & \emph{Points} & 62.0 & 77.9 & \textbf{95.9} & 68.4 & 69.0 \\%JointSSG_full_l20_6
    % & \emph{Points} & 66.5 & 80.6 & \textbf{95.5} & 77.4 & 71.5 \\%JointSSG_full_l20_6_1
    & $\sRotationFeature_{i\to j}$ & \textbf{66.1} & \textbf{81.2} & 95.6 & \textbf{77.4} & \textbf{71.5}\\%JointSSG_full_l20_5
    \midrule
    % \multirow{2}{*}{\rotatebox{90}{\emph{Dense}}} 
    \multirow{2}{*}{\emph{Dense}} 
    % &  & 32.9 & 55.5 & 89.1 & 41.0 & 31.4\\%JointSSG_inseg_l20_5
    & \emph{Points} & \textbf{35.1} & \textbf{57.5} & 89.6 & \textbf{47.9} & 33.2\\%  JointSSG_inseg_l20_6
    & $\sRotationFeature_{i\to j}$ & 34.1 & 58.1 & \textbf{89.9} & 43.0 & \textbf{33.3}\\% JointSSG_inseg_l20_1
    \midrule
    % \multirow{2}{*}{\rotatebox{90}{\emph{Sparse}}} 
    \multirow{2}{*}{\emph{Sparse}} 
    & \emph{Points} & 10.0 & 28.7 & \textbf{90.6} & 21.1 & 16.3\\%  JointSSG_orbslam_l20_11_9
    & $\sRotationFeature_{i\to j}$ & \textbf{9.9} & \textbf{29.5} & 90.4 & \textbf{23.5} & \textbf{16.5}\\%JointSSG_orbslam_l20_11_4
    \bottomrule
    \end{tabular}
    \caption{Ablation study on the use of input type for computing edge feature. The experiment setup is the same as in Tbl.~1. Here \emph{Points} means taking the point cloud union as in \cite{Wald2020_3dssg}, and $\sRotationFeature_{i\to j}$ is the relative pose descriptor described in Sec.~3.2.2.}
    \label{tbl:abla_edge_feature}
\end{table}

\subsection{Gate on the geometric feature}
whether to use the gate in the geometric feature or to use different ways of selecting keyframes.
We provide an additional ablation study on the effect of using a sigmoid gate when including the geometric feature in our message passing payer.

\begin{table}[t]
    \centering
    \begin{tabular}{c|c|rrr|rr}
    \toprule
    \multicolumn{1}{c}{~} & \multicolumn{1}{c}{~} & \multicolumn{3}{c}{\emph{Recall}(\%)} & \multicolumn{2}{c}{\emph{mRecall}}\\
    \multicolumn{1}{c}{~} & \multicolumn{1}{c}{Gate} & Rel. & Obj. & Pred. & Obj. & Pred. \\
    \midrule
    % \multirow{2}{*}{\rotatebox{90}{\emph{GT}}} 
    \multirow{2}{*}{\emph{GT}}
    &  & 61.5 & 77.1 & 95.3 & 77.1 & 70.9 \\%JointSSG_full_l20_5_4
    & \checkmark & \textbf{66.1} & \textbf{81.2} & \textbf{95.6} & \textbf{77.4} & \textbf{71.5}\\%JointSSG_full_l20_5
    \midrule
    % \multirow{2}{*}{\rotatebox{90}{\emph{Dense}}} 
    \multirow{2}{*}{\emph{Dense}} 
    &  & 32.9 & 55.5 & 89.1 & 41.0 & 31.4\\%JointSSG_inseg_l20_5
    & \checkmark & \textbf{34.1} & \textbf{58.1} & \textbf{89.9} & \textbf{43.0} & \textbf{33.3}\\% JointSSG_inseg_l20_1
    \midrule
    % \multirow{2}{*}{\rotatebox{90}{\emph{Sparse}}} 
    \multirow{2}{*}{\emph{Sparse}} 
    & & 8.6 & 26.9 & \textbf{90.5} & 24.4 & 15.6 \\%JointSSG_orbslam_l20_11_3
    & \checkmark & \textbf{9.9} & \textbf{29.5} & 90.4 & \textbf{23.5} & \textbf{16.5}\\%JointSSG_orbslam_l20_11_4
    \bottomrule
    \end{tabular}
    \caption{Ablation study on the proposed gated geometric feature. We ablate the use of a sigmoid function of a gate for the input geometric feature using the same experimental setup as in Tbl.~1.}
    \label{tbl:abla_geo_gate}
\end{table}

\section{Additional Results}

\subsection{Per-class prediction result}
In \TBL{ssg_3RScan_ScanNet20_PerClass}, We provide the per entity class recall for the experiment reported in Tbl.~1. Our method has dominant performance on most of the classes regardless of the input segmentation types.

\begin{table*}
    \centering
    \resizebox{\textwidth}{!}{
    \begin{tabular}{c|c|c|c|c|c|c|c|c|c|c|c|c|c|c|c|c|c|c|c|c|c|c}
         & &  bath. & bed & bkshf & cab. & chair & cntr. & curt. & desk & door & floor & ofurn & pic. & refri. & show. & sink & sofa & table & toil. & wall & wind. & avg \\
\hline
\multirow{5}{*}{\rotatebox{90}{\emph{GT}}} 
&IMP&
0.000 & 0.667 & 0.143 & 0.562 & 0.688 & 0.677 & 0.712 & 0.292 & 0.541 & 0.957 & 0.262 & 0.727 & 0.000 & 0.143 & 0.617 & 0.579 & 0.731 & 0.889 & 0.856 & 0.560 & 0.530\\%IMP_full_l20_6
&VGfM&
0.750 & 0.833 & \textbf{0.286} & 0.423 & 0.854 & 0.677 & 0.712 & 0.042 & 0.658 & \textbf{0.969} & 0.319 & 0.693 & 0.000 & 0.286 & 0.633 & 0.592 & 0.641 & 0.852 & 0.847 & 0.429 & 0.575\\%VGfM_full_l20_8
&3DSSG&
0.500 & 0.333 & 0.000 & 0.477 & 0.838 & 0.516 & 0.432 & 0.417 & 0.622 & 0.957 & 0.288 & 0.205 & 0.000 & 0.143 & 0.717 & 0.605 & 0.563 & 0.630 & 0.609 & 0.357 & 0.460\\%3DSSG_full_l20_3
&SGFN&
1.000 & 0.833 & 0.143 & 0.385 & 0.696 & \textbf{0.839} & 0.577 & 0.417 & 0.631 & 0.963 & 0.372 & 0.830 & 0.111 & 0.143 & 0.783 & 0.434 & 0.647 & 0.593 & 0.718 & 0.429 & 0.577\\%SGFN_full_l20_2
% &2DSSG&
% 1.000 & 1.000 & 1.000 & 0.500 & 0.947 & 0.774 & 0.919 & 0.500 & 0.829 & 0.975 & 0.372 & 0.875 & 1.000 & 1.000 & 0.950 & 0.737 & 0.737 & 1.000 & 0.799 & 0.643 & 0.828\\%2DSSG_full_l20_4
&Ours&
\textbf{1.000} & \textbf{1.000} & 0.000 & \textbf{0.619} & \textbf{0.927} & 0.645 & \textbf{0.847} & \textbf{0.583} & \textbf{0.838} & \textbf{0.969} & \textbf{0.539} & \textbf{0.943} & \textbf{0.667} & \textbf{1.000} & \textbf{0.900} & \textbf{0.671} & \textbf{0.808} & \textbf{1.000} & \textbf{0.877} & \textbf{0.655} & \textbf{0.774}\\%JointSSG_full_l20_5
\hline
\multirow{5}{*}{\rotatebox{90}{\emph{Dense}}} 
&IMP&
0.000 & 0.667 & 0.000 & 0.381 & 0.453 & 0.000 & 0.477 & 0.000 & 0.081 & 0.951 & 0.199 & 0.023 & 0.000 & 0.000 & 0.200 & 0.474 & 0.485 & \textbf{0.667} & 0.770 & \textbf{0.179} & 0.300\\%IMP_INSEG_l20_3
&VGfM&
0.000 & 0.667 & 0.000 & 0.346 & 0.494 & 0.000 & 0.486 & 0.042 & 0.198 & 0.957 & 0.141 & 0.011 & 0.000 & 0.000 & 0.233 & \textbf{0.579} & 0.569 & 0.630 & \textbf{0.780} & \textbf{0.179} & 0.316\\%VGfM_inseg_l20_1
&3DSSG&
0.250 & 0.667 & 0.000 & 0.200 & 0.510 & \textbf{0.258} & \textbf{0.505} & 0.000 & \textbf{0.477} & 0.914 & 0.147 & 0.034 & \textbf{0.222} & \textbf{0.143} & 0.250 & 0.474 & 0.425 & 0.259 & 0.519 & 0.131 & 0.319\\%3DSSG_INSEG_l20_3
&SGFN&
\textbf{0.750} & 0.333 & 0.000 & \textbf{0.508} & 0.636 & 0.194 & 0.405 & 0.083 & 0.387 & \textbf{0.969} & 0.230 & \textbf{0.114} & 0.111 & 0.000 & \textbf{0.383} & 0.553 & \textbf{0.623} & 0.519 & 0.730 & 0.131 & 0.383\\%SGFN_inseg_l20_0
% &2DSSG&
% 1.000 & 1.000 & 0.143 & 0.604 & 0.632 & 0.194 & 0.550 & 0.000 & 0.468 & 0.957 & 0.194 & 0.125 & 0.000 & 0.286 & 0.383 & 0.566 & 0.629 & 0.741 & 0.651 & 0.262 & 0.469\\%2DSSG_inseg_l20_0
&Ours&
\textbf{0.750} & \textbf{1.000} & 0.000 & 0.504 & \textbf{0.656} & 0.194 & 0.459 & \textbf{0.125} & 0.342 & \textbf{0.969} & \textbf{0.251} & 0.057 & 0.000 & \textbf{0.143} & \textbf{0.383} & \textbf{0.579} & 0.599 & \textbf{0.667} & 0.761 & 0.155 & \textbf{0.430}\\%JointSSG_inseg_l20_1
\hline
\multirow{5}{*}{\rotatebox{90}{\emph{Sparse}}} 
 & IMP & 
 0.000 & \textbf{0.333} & 0.000 & 0.235 & 0.146 & \textbf{0.129} & 0.252 & \textbf{0.167} & 0.099 & 0.798 & 0.068 & 0.023 & 0.000 & \textbf{0.286} & 0.183 & \textbf{0.329} & \textbf{0.281} & 0.259 & 0.324 & \textbf{0.214} & 0.206\\%IMP_orbslam_l20_1
 & VGfM & 
 0.000 & \textbf{0.333} & 0.000 & \textbf{0.273} & 0.150 & 0.097 & 0.243 & 0.042 & 0.081 & \textbf{0.810} & 0.047 & 0.011 & 0.000 & 0.000 & 0.150 & 0.276 & 0.263 & 0.222 & 0.325 & 0.202 & 0.176\\%VGfM_orbslam_l20_1
 & 3DSSG & 
 0.000 & 0.000 & 0.000 & 0.085 & 0.045 & 0.000 & 0.108 & 0.000 & 0.063 & 0.558 & 0.005 & 0.011 & 0.000 & 0.000 & 0.017 & 0.000 & 0.186 & 0.000 & 0.048 & 0.060 & 0.059\\%3DSSG_3rscan_orbslam_l20_1
 & SGFN & 
 0.000 & 0.000 & 0.000 & 0.058 & 0.081 & 0.032 & 0.072 & 0.042 & 0.063 & 0.712 & 0.010 & 0.034 & 0.000 & 0.000 & 0.083 & 0.092 & 0.174 & 0.000 & 0.097 & 0.107 & 0.083\\%SGFN_3rscan_orbslam_l20_1
 % & 2DSSG & 
 % 0.500 & 0.500 & 0.143 & 0.250 & 0.231 & 0.226 & 0.396 & 0.208 & 0.198 & 0.681 & 0.094 & 0.068 & 0.000 & 0.143 & 0.233 & 0.263 & 0.299 & 0.370 & 0.264 & 0.214 & 0.264\\%2DSSG_orbslam_l20_0
 & Ours & 
 \textbf{0.500} & \textbf{0.333} & \textbf{0.286} & \textbf{0.273} & \textbf{0.227} & \textbf{0.129} & \textbf{0.270} & 0.042 & \textbf{0.135} & \textbf{0.810} & \textbf{0.110} & \textbf{0.068} & 0.000 & 0.000 & \textbf{0.217} & 0.211 & 0.263 & \textbf{0.296} & \textbf{0.342} & 0.179 & \textbf{0.235}\\%Joint_orbslam_l20_11_4
    \end{tabular}
    }
    \caption{The per-class \emph{Recall} of all methods in 3RScan dataset~\cite{Wald2019RIO} with 20 node classes.}
    \label{tbl:ssg_3RScan_ScanNet20_PerClass}
\end{table*}

\subsection{Without consider \emph{None} estimation}
Our evaluations follow the line of work~\cite{Wald2020_3dssg,wu2021scenegraphfusion} which consider \emph{None} relationship is crucial, unlike other work~\cite{Xu2017,Gay2018_visual} which only consider edges with annotated non-\emph{None} relationships. Both approaches have their advantages. Considering \emph{None} estimation prevents excessive relationship estimation while also preventing potential relationship discovery, \eg should exist but was not annotated. For further comparison and the interest of potential readers, we provide the evaluation result without considering the \emph{None} relationship in \cref{tbl:3dssg_full_eval_l20_wo_none} and \cref{tbl:3dssg_full_l160_wo_none}, with the experiment setup as reported in Tbl.~1 and Tbl.~2.

\setlength{\tabcolsep}{4.0pt}
\begin{table}[t]
\begin{center}%
\begin{tabular}{c|l|rrr|rr}%
\toprule%
\multicolumn{1}{c}{~} & \multicolumn{1}{c}{\multirow{2}{*}{Method}} & \multicolumn{3}{c}{\emph{Recall}(\%)} & \multicolumn{2}{c}{\emph{mRecall}(\%)} \\
\multicolumn{1}{c}{~} & \multicolumn{1}{c}{~}  &  \multicolumn{1}{c}{Rel.} & Obj. & \multicolumn{1}{c}{Pred.} & \multicolumn{1}{c}{Obj.} & Pred.\\
\midrule%
\multirow{5}{*}{\rotatebox{90}{\emph{GT}}} 
        & IMP~\cite{Xu2017} & 43.5 & 70.1 & 54.2 & 53.0 & 38.1\\%IMP_full_l20_6
        & VGfM~\cite{Gay2018_visual} & 49.0 & 69.4 & 61.9 & 57.5 & 44.6\\%VGfM_full_l20_8
        & 3DSSG~\cite{Wald2020_3dssg} & 40.4 & 58.0 & 69.3 & 46.8 & 58.7\\%3DSSG_full_l20_3
        & SGFN~\cite{wu2021scenegraphfusion} & 47.2 & 63.8 & 71.4 & 57.7 & 65.5\\%SGFN_full_l20_2
        % & 2DSSG & 61.5 & 76.3 & 79.2 & 82.8 & 75.2\\%2DSSG_full_l20_4
        & Ours & \textbf{64.8} &\textbf{ 81.2} & \textbf{76.8} & \textbf{77.4} & \textbf{71.5}\\%JointSSG_full_l20_5
        \midrule
\multirow{5}{*}{\rotatebox{90}{\emph{Dense}}} 
        & IMP~\cite{Xu2017} & 18.3 & 51.8 & 19.3 & 30.0 & 23.0\\%IMP_INSEG_l20_3
        & VGfM~\cite{Gay2018_visual} & 20.8 & 53.3 & 22.1 & 31.6 & 24.4\\%VGfM_inseg_l20_1
        & 3DSSG~\cite{Wald2020_3dssg} & 15.1 & 41.4 & 26.1 & 31.9 & 26.6\\%3DSSG_INSEG_l20_3
        & SGFN~\cite{wu2021scenegraphfusion}  & 24.4 & 56.7 & 27.2 & 38.3 & 30.5\\%SGFN_inseg_l20_0
        % & 2DSSG & 23.5 & 56.8 & 28.4 & 46.9 & 33.9\\%2DSSG_inseg_l20_0 | 2DSSG_inseg_l20_0
        & Ours & \textbf{25.5} & \textbf{58.1} & \textbf{27.3} & \textbf{43.0} & \textbf{33.3}\\%Joint_inseg_l20_1
        \midrule
\multirow{6}{*}{\rotatebox{90}{\emph{Sparse}}} 
        & IMP~\cite{Xu2017} & 3.5 & 27.5 & 4.0 & 20.6 & 14.0\\%IMP_orbslam_l20_1
        & VGfM~\cite{Gay2018_visual} & 3.7 & 26.9 & 4.2 & 17.6 & 15.4\\%VGfM_orbslam_l20_1
        & 3DSSG~\cite{Wald2020_3dssg} & 1.0 & 9.7 & \textbf{9.2} & 5.9 & 15.1\\%3DSSG_3rscan_orbslam_l20_1
        & SGFN~\cite{wu2021scenegraphfusion} & 2.3 & 12.6 & 7.0 & 8.3 & 14.4\\%SGFN_3rscan_orbslam_l20_1
        % & 2DSSG  & 5.8 & 27.5 & 10.1  & 26.4 & 19.7\\%2DSSG_orbslam_l20_0 %
        & Ours & 6.9 & 29.5 & 8.1 & 23.5 & \textbf{16.5}\\%JointSSG_orbslam_l20_11_4
        & Ours (i) & \textbf{7.1} & \textbf{30.2} & 8.4 & \textbf{24.5} & 15.9\\%JointSSG_orbslam_l20_11_4
\bottomrule
\end{tabular}
\end{center}
\caption{We report top-1 recall of method without considering \emph{None} relationship. The experiment setup is the same as in Tbl~1.}
\label{tbl:3dssg_full_eval_l20_wo_none} 
\end{table}%

%%%% Change to instance level eval.
%\begin{wraptable}{r}{7.5cm}
\begin{table}[t]
%\vspace{-0.8cm}
% \begin{table}[]
    \centering
% \resizebox{\linewidth}{!}{
    \begin{tabular}{l|ccc|cc}
    \toprule
      \multicolumn{1}{c}{\multirow{2}{*}{Method}}     & \multicolumn{3}{c}{\emph{Recall}(\%)}       & \multicolumn{2}{c}{\emph{mRecall}(\%)} \\
    \multicolumn{1}{c}{~}  &  \multicolumn{1}{c}{Rel.} & Obj. & \multicolumn{1}{c}{Pred.} & \multicolumn{1}{c}{Obj.} & Pred.\\
    \midrule
        IMP~\cite{Xu2017} & 3.3 & 35.9 & 9.0 & 18.7 & 4.9\\%IMP_FULL_l160_2_3
        VGfM~\cite{Gay2018_visual} & 3.4 & 37.9 & 14.7 & 17.9 & 6.5\\%VGfM_FULL_l160_3_1
        3DSSG~\cite{Wald2020_3dssg} & 7.3 & 29.6 & \textbf{68.8} & 11.7 & \textbf{25.5}\\%3DSSG_full_l160_1
        SGFN~\cite{wu2021scenegraphfusion} & 4.5 & 29.4 & 45.8 & 11.8 & 13.5\\%SGFN_full_l160_4
        % 2DSSG & 46.8 & 51.4 & 44.9 & 29.7 & 30.2\\ %2DSSG_full_l160_2
        Ours & \textbf{17.9} & \textbf{56.7} & 50.4 & \textbf{27.2} & 23.9\\%JointSSG_full_l160_0
        \bottomrule
    \end{tabular}
% }
\caption{We report top-1 recall of method without considering \emph{None} relationship. The experiment setup is the same as in Tbl~2.
}
\label{tbl:3dssg_full_l160_wo_none}
\vspace{-0.3cm}
\end{table}
%\end{wraptable}

%%%%%%%%% REFERENCES
{\small
\bibliographystyle{ieee_fullname}
\bibliography{egbib}
}